\documentclass{article}

\usepackage{microtype}
\usepackage{graphicx}
\usepackage{subcaption}
\usepackage{booktabs} 
\usepackage[T1]{fontenc}

\usepackage{hyperref}

\usepackage{amsmath}
\usepackage{amssymb}
\usepackage{mathtools}
\usepackage{amsthm}
\usepackage{multirow}
\usepackage{pifont}
\usepackage{enumitem}

\usepackage[capitalize,noabbrev]{cleveref}

\usepackage[preprint]{icml2026}

\theoremstyle{plain}

\theoremstyle{definition}

\theoremstyle{remark}

\usepackage{amsmath,amsfonts,bm}

\def\eqref#1{equation~\ref{#1}}

\def\1{\bm{1}}

\DeclareMathAlphabet{\mathsfit}{\encodingdefault}{\sfdefault}{m}{sl}
\SetMathAlphabet{\mathsfit}{bold}{\encodingdefault}{\sfdefault}{bx}{n}

\newcommand{\cmark}{\ding{51}}
\newcommand{\xmark}{\ding{55}}
\newcommand{\bQ}{b_Q}

\newcommand{\bV}{b_V}
\newcommand{\WQ}{W_Q}
\newcommand{\WK}{W_K}
\newcommand{\WV}{W_V}
\newcommand{\WO}{W_O}

\usepackage[textsize=tiny]{todonotes}

\icmltitlerunning{Symmetry-breaking attention bias alignment}

\begin{document}

\twocolumn[
  \icmltitle{Symmetry Breaking in Transformers for Efficient and Interpretable Training}




  \begin{icmlauthorlist}
    \icmlauthor{Eva Silverstein}{stan}
    \icmlauthor{Daniel Kunin}{ucb}
    \icmlauthor{Vasudev Shyam}{zyph}
  \end{icmlauthorlist}

  \icmlaffiliation{stan}{Stanford University}
  \icmlaffiliation{ucb}{UC Berkeley}
  \icmlaffiliation{zyph}{Zyphra Technologies}

  \icmlcorrespondingauthor{Eva Silverstein}{evas@stanford.edu}

  \icmlkeywords{Machine Learning, ICML}

  \vskip 0.3in
]



\printAffiliationsAndNotice{}  

\begin{abstract}
    The attention mechanism in its standard implementation contains extraneous rotational degrees of freedom that are carried through computation but do not affect model activations or outputs. We introduce a simple symmetry-breaking protocol that inserts a preferred direction into this rotational space through batchwise-sampled, unlearned query and value biases.
    This modification has two theoretically motivated and empirically validated consequences.
    First, it can substantially improve the performance of simple, memory-efficient optimizers, narrowing—and in some cases closing—the gap to successful but more complex memory-intensive adaptive methods. 
    We demonstrate this by pretraining 124M parameter transformer models with four optimization algorithms (AdamW, SOAP, SGDM, and Energy Conserving Descent(ECD)) and evaluating both validation loss and downstream logical reasoning.
    Second, it enables an interpretable use of otherwise redundant rotational degrees of freedom, selectively amplifying semantically meaningful token classes within individual attention heads. 
    Overall, our results show that minimal, principled architectural changes can simultaneously improve performance and interpretability.
\end{abstract}

\section{Introduction}

Transformers exhibit remarkable capabilities, and it is of interest to understand and improve their architecture and training in order to further accelerate progress in artificial intelligence.  In this work, we introduce a simple architectural modification—batchwise, untrained query and value biases—that facilitates more efficient optimization and improves model performance. We motivate this modification theoretically and evaluate it empirically by pretraining GPT-2 (124M) models, measuring both validation loss and downstream logical reasoning performance.  
   

In practice, transformer training has been dominated by adaptive or preconditioned variants of stochastic gradient descent with momentum (SGDM), such as AdamW \cite{KingmaBa2015, LoshchilovHutter2019} and SOAP \cite{Vyasetal2024}. These methods are typically memory intensive, requiring on the order of $3N$ auxiliary variables for an $N$-parameter model, and often rely on heuristic arguments to motivate how adaptivity is implemented. By contrast, the recently proposed Energy Conserving Descent (ECD) family of optimizers \cite{BBI, DeLuca:2023yld, DeLuca:2025ruv, robnik2023microcanonical} adopts a predictive physics-inspired framework grounded in  Hamiltonian dynamics and requires only $2N$ auxiliary variables. Despite these appealing properties—and demonstrated advantages in scientific applications—prior work has shown that ECD does not match the empirical performance of adaptive optimizers in transformer training.

In this work, we identify a reason for this disparity and develop a simple intervention that overcomes it while providing additional, interpretable benefits. Our analysis focuses on the symmetry structure of the transformer architecture and its interaction with learning dynamics, building on earlier perspectives on symmetry and optimization \cite{2020arXiv201204728K, tanaka2021noether}. The key architectural component of transformers—the attention head—is invariant under a large group of continuous rotational symmetries \cite{zhang2025beyond}. For each head, a joint rotation of the query and key matrices preserves attention scores, since these depend only on their inner products; an analogous symmetry appears in the value–output sector. Although these directions carry no gradient signal and leave activations unchanged, they nevertheless shape the optimization dynamics. 

We analyze how this symmetry interacts with optimization, showing in particular why it can obstruct Energy Conserving Descent. From a Hamiltonian perspective, the rotational invariance induces conserved angular momenta along these symmetry directions, which restricts the chaotic exploration in parameter space required for ECD to effectively proceed in descent directions. SGDM is similarly affected, though its dynamics has one fewer conservation law.

Motivated by this analysis, we introduce a batchwise symmetry-breaking protocol that injects fixed, untrained query and value biases, denoted $b_Q$ and $b_V$, thereby introducing a preferred direction in the otherwise degenerate rotational subspace. Beyond improving optimization, the query bias $b_Q$ enables a novel, interpretable mechanism for shaping model behavior: the model can amplify or suppress classes of tokens by learning strong alignment or anti-alignment between their key vectors $k = W_K x$ and the mean bias $\mathbb{E}[b_Q]$, yielding an exponential modulation of attention weights proportional to $e^{k \cdot b_Q}$. Since widely used open-source models such as Gemma, Llama, DeepSeek, and Qwen do not include bias terms in their attention heads, this intervention in pretraining can be used without sacrificing any standard training parameters.

Our empirical analysis shows that this explicit symmetry breaking substantially improves memory-efficient optimization methods, particularly ECD. Across validation loss and downstream logical reasoning tasks, ECD trained with our protocol becomes competitive with SOAP at the GPT-2 (124M) scale.  In our experiments, ECD exhibits consistent use of the query bias to amplify semantically meaningful token classes, a property which for all optimizers is predictive of improved downstream reasoning performance.  

Overall, our results highlight how a careful analysis of architectural symmetries and learning dynamics can reveal simple, principled modifications that improve both efficiency and interpretability. We leave to future work a broader evaluation across model scales and optimization schemes, including the integration of $b_Q, b_V$ symmetry breaking and ECD into the benchmarking protocol of \citet{wen2025fantastic}.

\textbf{Our contributions.}
We summarize our main theoretical and empirical contributions as follows:
\begin{enumerate}
    \item We provide a Hamiltonian explanation for why ECD fails to train transformer models out of the box, showing that rotational symmetries in attention induce conserved quantities that can obstruct descent (\cref{sec:symmetries-conservation-optimization}).
    \item We propose a simple, symmetry-breaking modification that lifts this obstruction while preserving the memory efficiency and structural simplicity of ECD (\cref{sec-rotation-breaking-bQ-bV}).
    \item We illustrate empirically that symmetry-broken ECD matches—and in some cases exceeds—the performance of adaptive optimizers such as Adam and SOAP on GPT-2--scale transformer models (\cref{sec:performance}).
    \item We show that the proposed symmetry-breaking mechanism is interpretable, enabling direct analysis of how it alters learning outcomes within attention layers to amplify or suppress token classes (\cref{sec:bq_alignment}).\footnote{Code available at \url{https://github.com/evasilverstein/Symmetry-breaking-attention-bias}}
\end{enumerate}

\section{Related Work}

Our work engages with several lines of research, including the optimization of transformer models and energy-based, physics-inspired optimization methods.
Additionally, our work provides a potential window into interpretability \cite{sharkey2025open} through a novel alignment mechanism.

\textbf{Optimization of transformer models.}
Most modern deep learning training pipelines rely on adaptive variants of stochastic gradient descent with momentum (SGDM) \cite{RobbinsMonro1951}, such as Adam \cite{KingmaBa2015} and AdamW \cite{LoshchilovHutter2019}. 
Several other methods have significantly influenced modern training practice, including Polyak’s heavy-ball momentum \cite{Polyak1964}, RMSprop \cite{TielemanHinton2012}, and the memory-efficient Adafactor \cite{ShazeerStern2018}.
To improve convergence through structured preconditioning, Shampoo \cite{Guptaetal2018} introduced second-order, structure-aware updates, which have since been optimized for large-scale language modeling via SOAP \cite{Vyasetal2024}.  
More recently, scalable alternatives such as Sophia \cite{Liuetal2024Sophia} and Muon \cite{Liuetal2025} have been proposed to balance adaptivity, stability, and memory efficiency in large-model training.
A consistent empirical observation across this literature is that adaptive optimizers tend to outperform simpler, memory-light methods such as SGDM when training transformer architectures, particularly at scale.
Our work aims to close this gap by introducing a simple architectural modification, guided by the geometric and symmetry structure of attention and its interaction with the dynamics of both adaptive and energy-based optimizers.

\textbf{Energy-based and Hamiltonian optimization.}
Energy Conserving Descent (ECD) and related physics-inspired methods have emerged as a promising alternative to friction-dominated optimization dynamics \cite{BBI, DeLuca:2023yld, DeLuca:2025ruv, robnik2023microcanonical}. 
Rather than relying on adaptive per-parameter statistics, these approaches formulate learning as a chaotic Hamiltonian dynamical system in which total energy is conserved and the framework predicts a formula (based on the Liouville measure) for distribution of training results as a function of the loss.
Variants of ECD have been applied to likelihood-free inference in particle physics \cite{DeLuca:2025ruv}, while Microcanonical Hamiltonian and Langevin Monte Carlo (MCHMC/MCLMC) methods have demonstrated advantages in cosmological field-level inference \cite{BayerSeljakModi2023} and Bayesian neural network sampling \cite{Sommeretal2025}. 
These results highlight the potential of energy-based dynamics in high-dimensional scientific and statistical settings. 
However, in large-scale transformer architectures these methods have not taken hold, and in our experiments consistently exhibited poor empirical performance.
We provide a mechanistic explanation for this failure in attention-based models and identify a principled architectural modification that restores effective optimization, with interpretable benefits regardless of optimizer.

\section{Symmetry, Conservation, and Optimization}
\label{sec:symmetries-conservation-optimization}

\emph{Why does ECD fail to match the empirical performance of adaptive optimizers at training transformers?} In this section, we argue that the answer lies in the interaction between the symmetry structure of the transformer architecture and the conservation laws that govern the dynamics of ECD. 

To make this connection precise, we first introduce a unifying Hamiltonian perspective on SGDM and a simple version of ECD, which makes explicit how continuous symmetries give rise to conserved quantities. We then apply this framework to the attention mechanism in transformers, showing that its large group of continuous rotational symmetries induces conserved angular momenta that are particularly constraining for ECD's energy-conserving, chaotic Hamiltonian dynamics. This analysis motivates the symmetry-breaking mechanism we propose in \S \ref{sec-rotation-breaking-bQ-bV}. Readers most interested in this mechanism may proceed directly to \S\ref{sec:transformer-rotations} and then \S\ref{sec-rotation-breaking-bQ-bV}.

\subsection{A Hamiltonian Framework for Optimization}

Hamiltonian mechanics provides a natural unifying framework for studying optimization algorithms (see \citet{wibisono2016variational} for related work).
Our goal in this subsection is to make precise how SGDM and ECD arise as different choices within a family of dynamical systems.

Let $\Theta \in \mathbb{R}^N$ denote the parameters of a neural network and $\Pi \in \mathbb{R}^N$ their conjugate momenta, which can be understood as a velocity-like state that accumulates past gradients. Consider the evolution of $(\Theta, \Pi)$ under Hamilton’s equations,
\begin{equation}
\label{eq:hamiltons-eqs}
\dot{\Theta} = \frac{\partial H}{\partial \Pi},
\qquad
\dot{\Pi} = -\frac{\partial H}{\partial \Theta},
\end{equation}
where overdots denote derivatives with respect to a continuous-time variable.
A wide class of optimization algorithms can be obtained by discretizing these equations with different choices of a Hamiltonian $H$ of the form
\begin{equation}
\label{eq-unified-Hamiltonian}
H(\Theta, \Pi, t)
=
e^{-f t} \frac{\|\Pi\|^2}{2 m(\Theta)}
+
e^{f t} V(\Theta).
\end{equation}
Here $V(\Theta)$ plays the role of a potential, $m(\Theta)$ is a (possibly position-dependent) mass, and $f \ge 0$ controls dissipation.
Selecting $V$ and $m$ as functions of the loss function $F(\Theta)$ being optimized and discretizing the resulting equations of motion yields standard friction-based optimizers as well as energy-conserving variants within a single framework. Importantly, the Hamiltonian $H(\Theta, \Pi, t)$ represents the total energy of the system. This energy may be time-dependent or time-independent. As we will see, the dissipative, time-dependent case corresponds to SGDM, while the time-independent, energy-conserving case gives ECD.

\begin{figure}[t]
    \centering
    \begin{subfigure}[t]{0.48\linewidth}
        \centering
        \includegraphics[width=\linewidth]{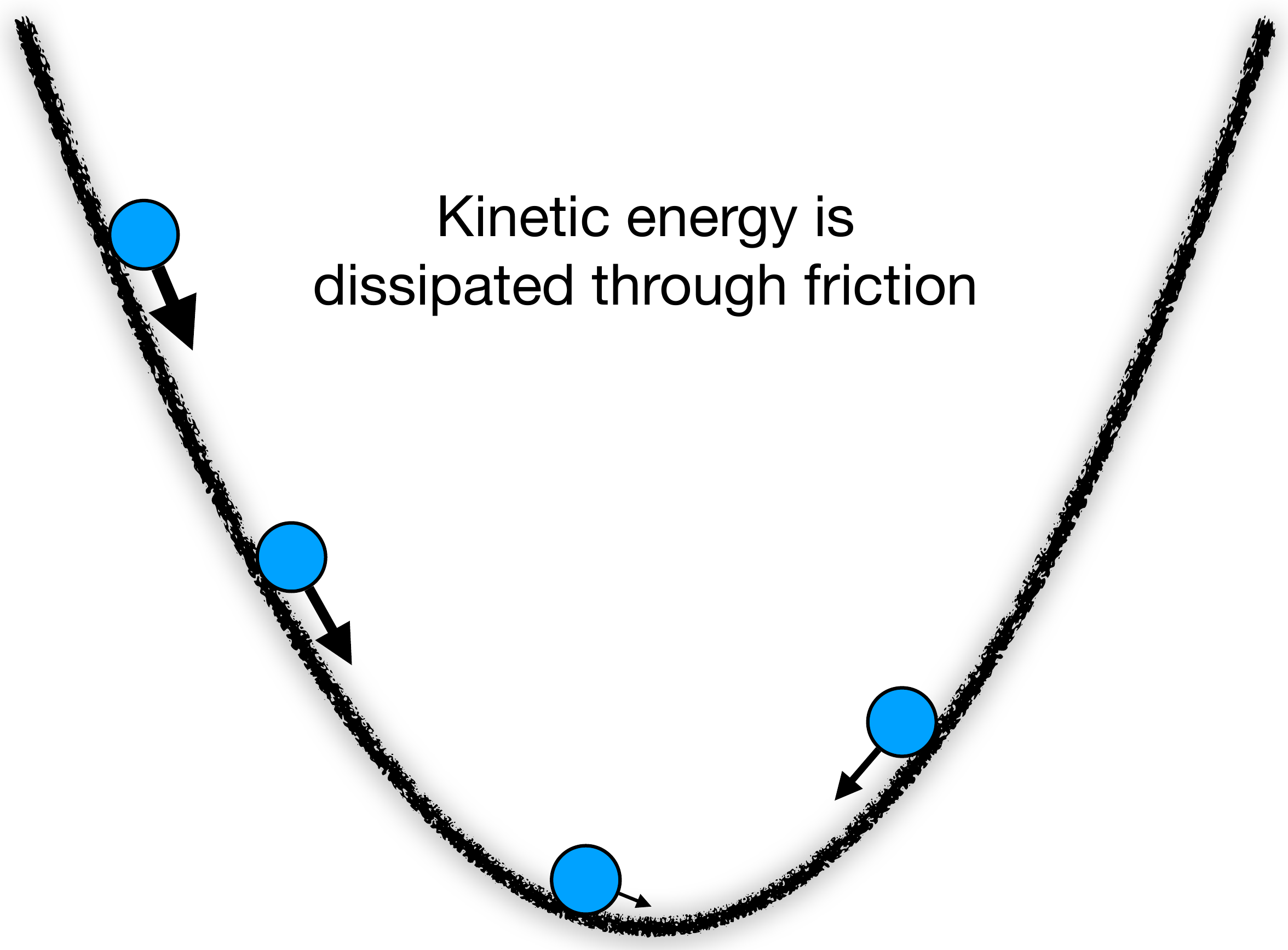}
        \caption{Friction-based (SGDM)}
        \label{fig:illustration-sgdm}
    \end{subfigure}\hfill
    \begin{subfigure}[t]{0.48\linewidth}
        \centering
        \includegraphics[width=\linewidth]{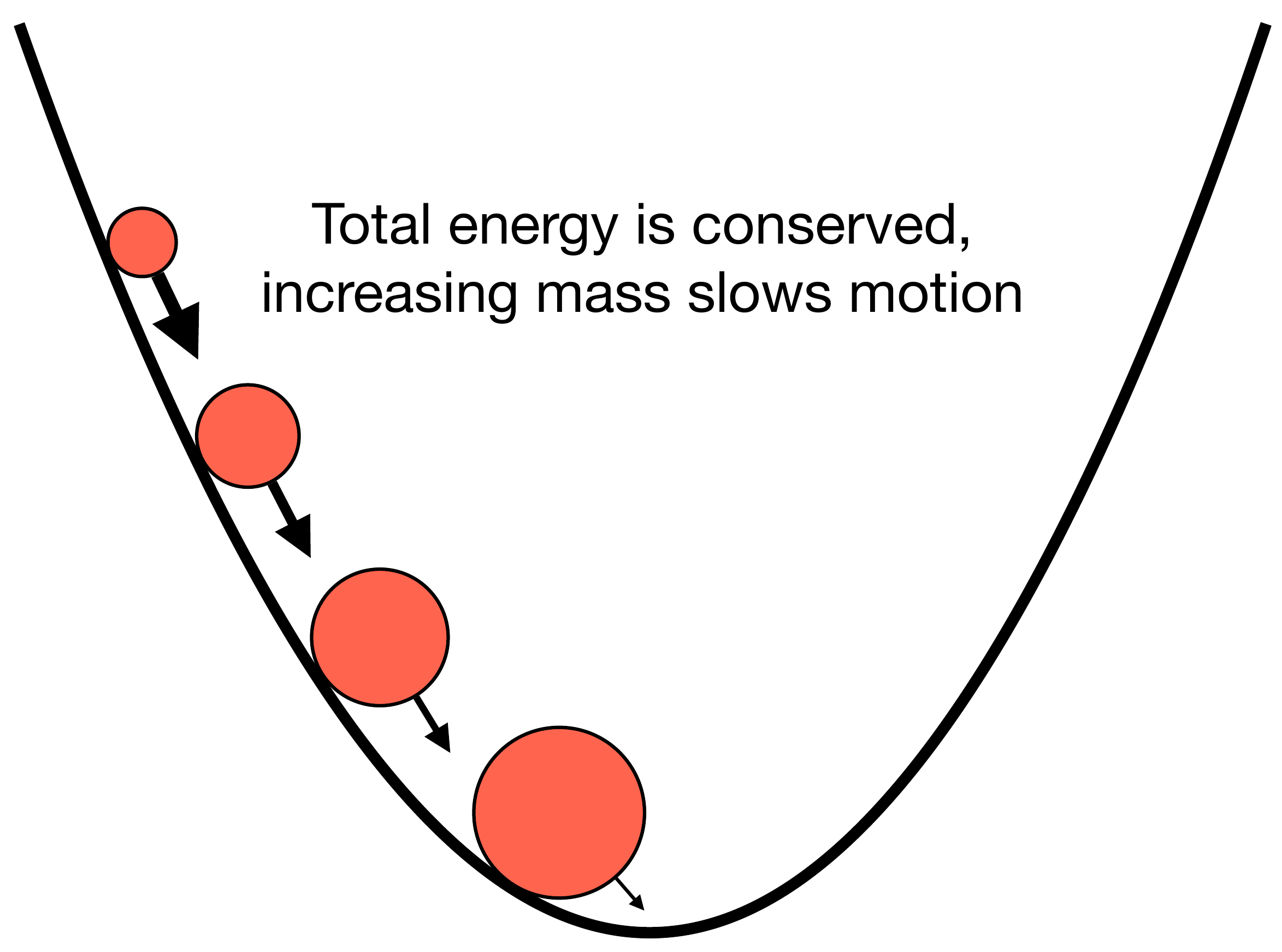}
        \caption{Energy-conserving (ECD)}
        \label{fig:illustration-ecd}
    \end{subfigure}
    \caption{\textbf{Friction-based versus energy-conserving optimization.} 
    Here we illustrate the difference between a friction-based versus energy-conserving optimization strategy in a quadratic basin.
    \textbf{(a)} SGDM is like a ball rolling downhill with friction: momentum can cause it to overshoot the minimum, but kinetic energy is dissipated, damping oscillations, leading to convergence. 
    \textbf{(b)} ECD is like as a snowball rolling in the same basin: here the total energy is conserved, but the mass increases as the loss decreases, naturally reducing velocity and slowing motion near the minimum.}
    \label{fig:illustration}
\end{figure}

\noindent{\bf Stochastic Gradient Descent with Momentum (SGDM)} corresponds to a dissipative Hamiltonian system obtained by setting the friction coefficient $f > 0$, keeping the mass $m$ constant, and identifying the potential with the loss, $V(\Theta) = F(\Theta)$. In continuous time, the resulting equations of motion take the form
\begin{equation}
    \dot{\Theta} = e^{-ft}\frac{\Pi}{m},
    \qquad
    \dot{\Pi} = -e^{ft} \nabla_\Theta F(\Theta).
\end{equation}
Taking a time derivative of the first equation here and substituting in the second yields the familiar second-order equation describing the motion of a particle driven by the gradient of the loss and slowed by a frictional force,
\begin{equation}
    \ddot{\Theta} = - f \dot{\Theta} - \nabla F(\Theta)/m.
\end{equation}
SGDM is a discretization of this equation, with additional stochasticity arising from minibatch estimates of the gradient.
The dynamics reduce to gradient flow, $\dot{\Theta} \propto - \nabla F(\Theta)$, in the friction-dominated limit, and to Newton's second law, $\ddot{\Theta} \propto - \nabla F(\Theta)$, in the zero-friction limit.

From this perspective, the role of friction is to regulate the speed of motion in parameter space: it damps oscillations
and progressively removes kinetic energy as the loss decreases. The total energy is not conserved, and convergence is enforced by continually draining velocity.

Adaptive descendants of SGDM, such as Adam and SOAP, extend this dissipative method by introducing an additional vector of internal state beyond $(\Theta, \Pi)$, effectively operating in a $3N$-dimensional space. Their adaptivity can be interpreted as state-dependent damping and preconditioning, which further shapes and slows motion in parameter space at the cost of increased memory and complexity.


\noindent{\bf Energy Conserving Descent (ECD)} adopts a fundamentally different perspective from dissipative methods by removing friction entirely.  A basic version of this idea simply reinterprets the loss landscape through a variable-mass Hamiltonian, as follows. Rather than treating the loss as a potential, ECD defines the effective mass as a function of the error, so that motion in parameter space slows as lower-loss regions are approached, (\cref{fig:illustration}). In its simplest form, this is achieved by removing friction $f = 0$, setting $V(\Theta)=0$, and defining $m(\Theta) = \big(F(\Theta) - F_0\big)^{-\eta}$, where $\eta>0$ controls the degree of concentration near low loss and $F_0$ acts as a prior on the expected minimum \cite{DeLuca:2025ruv}.
The resulting equations of motion, which follow from (\ref{eq:hamiltons-eqs}-\ref{eq-unified-Hamiltonian}), are
\begin{equation}
    \dot{\Theta} = \frac{\Pi}{m(\Theta)}, 
    \qquad 
    \dot{\Pi} = -\frac{E}{m(\Theta)}\, \frac{\partial m }{\partial \Theta},
    \label{eq:variable-mass-eoms}
\end{equation}
where $E = \frac{1}{2} m(\Theta)\, \dot{\Theta}^2$ is the total energy.
In contrast to SGDM, ECD operates in a time-independent Hamiltonian system: there is no friction and no explicit time dependence, and thus the total energy $E$ is conserved. 
This conservation law implies the induced Liouville measure on phase space, which yields a distribution over parameters of the form
\begin{equation}
    p(\Theta) \propto \big(F(\Theta) - F_0\big)^{-\frac{\eta d}{2}},
    \label{eq:liouville}
\end{equation}
governing the statistical distribution of results following a chaotic mixing time \cite{BBI, DeLuca:2023yld, DeLuca:2025ruv}.
This increasingly concentrates probability near low-loss regions as $\eta$ grows. This provides ECD with a novel mechanism for biasing exploration toward good solutions, not requiring the dissipative dynamics of friction-based methods \cite{BBI, DeLuca:2023yld, DeLuca:2025ruv, robnik2023microcanonical}.
In this work, we use a closely related formulation of the ECD framework that has proven effective in precision scientific applications \cite{DeLuca:2025ruv}, that we present in App. \ref{app:ECD-variants}.

\textbf{Noether’s theorem.}
The Hamiltonian formulation introduced above does more than unify SGDM and ECD under a common dynamical framework: it grants access to a powerful set of tools from classical mechanics for reasoning about optimization dynamics. Central among these is the connection between continuous symmetries and conserved quantities, formalized by Noether’s theorem \cite{Noether1918}.

If the Hamiltonian does not depend on some direction $\Theta_{\mathrm{symm}}$ in parameter space, then the conjugate momentum in that direction is conserved. From Hamilton’s equations,
\begin{equation}
\label{eq-Noether-momenta}
\dot{\Pi}_{\mathrm{symm}} = -\partial_{\Theta_{\mathrm{symm}}} H = 0,
\end{equation}
so invariance of $H$ under continuous translations along $\Theta_{\mathrm{symm}}$ implies conservation of the corresponding generator. This is the Hamiltonian version of Noether’s principle: every continuous symmetry leads to a conserved quantity.

\subsection{Continuous Rotational Symmetries in Attention}
\label{sec:transformer-rotations}  

The Hamiltonian framework for optimization highlights the central role of continuous symmetries in shaping learning dynamics. In deep learning, such symmetries can arise naturally from the structure of network parameterizations and can give rise, via Noether-type arguments, to conserved quantities that constrain motion in parameter space. Here, we study a central instance of this phenomenon: continuous rotational symmetries in the attention heads of transformers.

The attention heads of a transformer process the incoming residual stream $x$ according to
\begin{equation}\label{eq-attention}
W_O \sum_{j=1}^L 
\frac{m_{ij}\exp(x_i^T W_Q^T W_K x_j)}
{\sum_{m=1}^L m_{im}\exp(x_i^T W_Q^T W_K x_m)}
\, (W_V x_j),
\end{equation}
with trainable parameters in the weight matrices $W_Q, W_K, W_V,$ and $W_O$. Here $i$ indexes the query position, $j$ and $m$ index key positions, and $m_{ij} \in \{0,1\}$ accounts for causal or other masking.  For layer 0 attention, $x$ is simply embedded tokens.  

It is straightforward to verify that this formulation exhibits continuous rotational symmetries \cite{zhang2025beyond} in both the key--query and value--output weights.  
The key-query weights are invariant under the joint transformation
\begin{equation}\label{eq-rotations-Ws}
W_Q \rightarrow R W_Q, 
\qquad 
W_K \rightarrow R W_K,
\end{equation}
for any orthogonal matrix $R \in O(d_k)$, since the similarity scores depend only on the inner products $q_i \cdot k_j$.  
An analogous $O(d_v)$ rotational symmetry holds for the value--output mapping.
These symmetries are respected not only by the forward computation, but also by the Hamiltonian governing the learning dynamics of SGDM and ECD.
Each rotation group $O(d_{\text{head}})$ is generated by antisymmetric matrices $\Omega$ satisfying $\Omega^T = -\Omega$, yielding $d_{\text{head}}(d_{\text{head}}-1)/2$ independent generators.
By Noether’s theorem, these symmetries therefore give rise to conserved angular momenta in parameter space, corresponding to motion along the orbits of the associated symmetry groups.
With two independent rotation groups (Q--K and V--O) per attention head, the associated conserved Noether currents take the form
\begin{equation}
    \label{eq-Js-QK}
    J_\Omega^{QK} = m(\Theta)\left(\dot W_Q^T \Omega W_Q + \dot W_K^T \Omega W_K\right),
\end{equation}
and analogously for $J_\Omega^{VO}$.

\subsection{Implications of Conserved Angular Momentum}
\label{sec-ECD-SGDM-conserved-angular-momenta}

The rotational symmetries described above act independently within each attention head and each layer, so the full symmetry group of the model factorizes as $G_R = \big(O(d_k) \times O(d_v)\big)^{N_{\mathrm{attn}}}$, where $N_{\mathrm{attn}}$ denotes the total number of attention heads.  
For the GPT-2 124M model studied in this work, this group spans 580{,}608 parameters—only $\sim 0.5\%$ of the total parameter count—yet, as we now show, these symmetry directions play a disproportionate role in shaping optimization dynamics, in keeping with the importance of attention within the larger architecture.  
In particular, when the associated angular momenta are conserved, they can hinder ECD optimization in two closely related ways.

\begin{figure}[ht]
    \centering
    \includegraphics[width=0.49\textwidth]{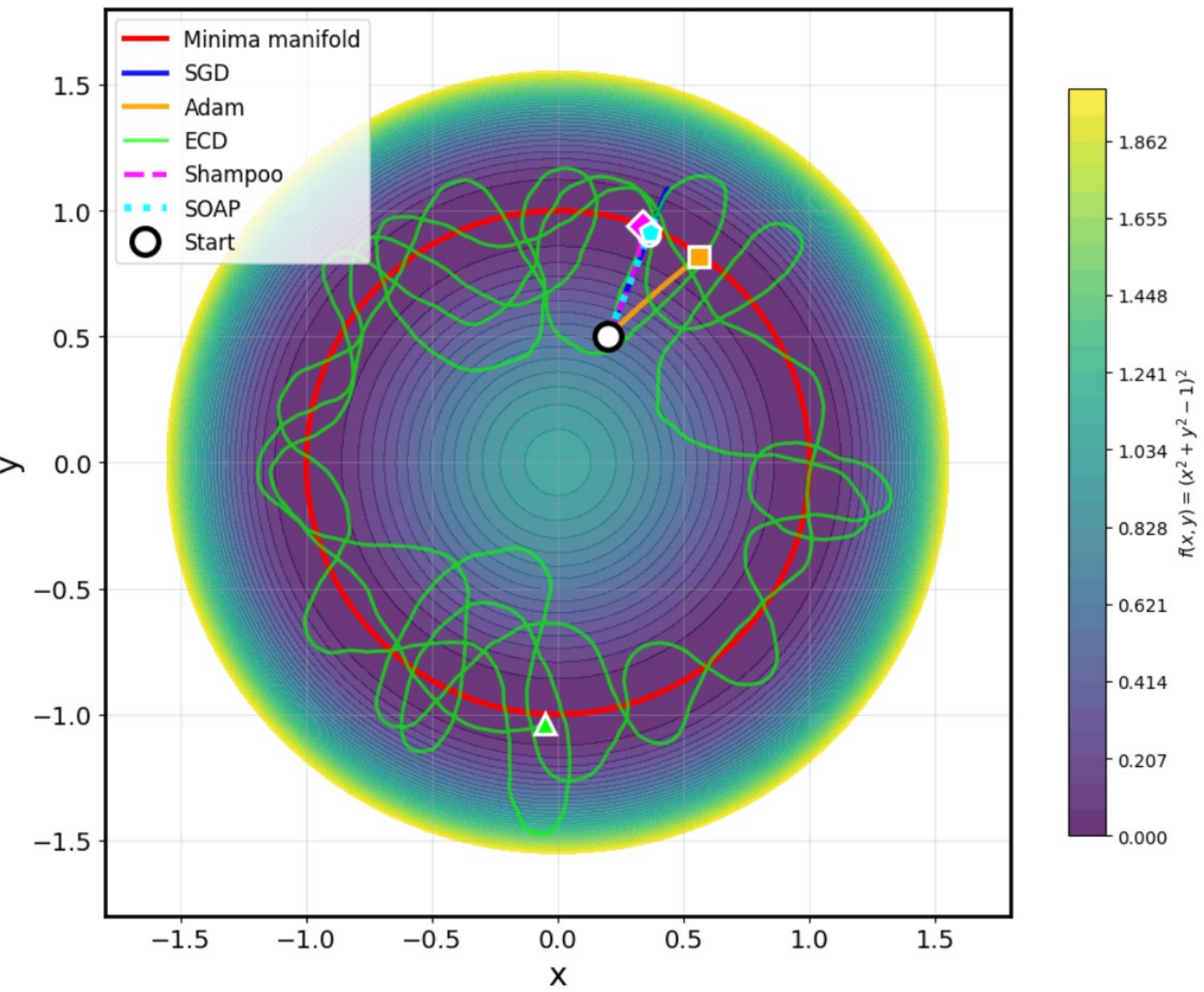}
    \caption{\textbf{Conserved angular momentum hinders ECD.}
    Trajectories of five optimizers on the synthetic error surface $f(x,y) = (x^2 + y^2 - 1 + z)^2$, which exhibits a circular symmetry and serves as a toy model for the rotational symmetries in attention.
    Random perturbations $z \sim \mathcal{N}(0, 0.1^2)$ mimic mini-batch noise.
    SGDM is initialized with zero momentum and consequently conserves zero angular momentum, proceeding radially toward the minimum.
    Adam with its fixed, symmetry-breaking coordinate axes proceeds along a different angle than SOAP and Shampoo, whose preconditioning steps orient them in a radial-angular bases.
    In contrast, ECD (shown with an exaggerated chaos parameter $\nu$) introduces angular momentum that interferes with optimization as in \S\ref{sec-ECD-SGDM-conserved-angular-momenta}.
    }
    \label{fig:sombrero}
\end{figure}

First, by diverting kinetic energy into rotational motion within the symmetry subspaces; because the total energy is fixed, this limits the available velocity in the remaining directions of parameter space, including those aligned with meaningful descent of the loss.
For a single attention head, decompose the parameters into tangential directions $\Theta_{\parallel}$ that spans the directions generated by $\Omega$ and orthogonal components $\Theta_{\perp}$ that span all other directions.
%
%
Using a relationship analogous to the relation between angular momentum, angular velocity, and the radius in classical mechanics, we can then schematically express the conserved energy $E = \frac{m(\Theta)}{2}\|\dot\Theta\|^2$ as,
\begin{equation}
E \;\sim\; \frac{m(\Theta)}{2}\|\dot{\Theta}_{\perp}\|^2 \;+\; \frac{1}{2m(\Theta)}\frac{\|J_\Omega\|^2}{\|\Theta_{\perp}\|^2},
\end{equation}
where $||J_\Omega||^2 = ||J_{\Omega}^{QK}||^2 + ||J_{\Omega}^{VO}||^2$. 
This makes explicit how, at fixed energy, nonzero angular momentum leads to a trade-off: increased motion along the symmetry orbits necessarily suppresses velocity in the orthogonal, loss-reducing directions, as depicted in \cref{fig:sombrero}.

Second, the presence of many conserved quantities suppresses chaotic mixing, undermining the assumptions behind using the Liouville measure (\ref{eq:liouville}) to characterize exploration in parameter space.  
Although ECD is initialized with velocity aligned to the gradient, so that angular momenta vanish initially, in practice, sufficiently fast mixing often requires introducing stochasticity in the momentum direction.  
Following \cite{robnik2023microcanonical, DeLuca:2023yld, DeLuca:2025ruv}, this is implemented in the version of ECD we use \cite{DeLuca:2025ruv}  through a small random rotation controlled by the hyperparameter $\nu$: when $\nu \neq 0$, angular momentum accumulates and further constrains motion in loss-reducing directions, while when $\nu = 0$, the dynamics may fail to mix adequately.  
Additional numerical effects can also introduce slow drift in angular momentum (see App. \ref{app:rotational_drift}).

In this section we argued that rotational symmetries in transformer attention heads induce conserved angular momenta that hinder loss-reducing dynamics. The next section shows how a minimal architectural change breaks these symmetries and restores efficient learning.

\section{Biases to break attention symmetries}
\label{sec-rotation-breaking-bQ-bV}

We are theoretically motivated to break the rotational symmetry in transformers.  In this section we devise a method to do this with unlearned bias terms, both to introduce a sufficiently chaotic landscape for exploration without conserved angular momenta, and to allow the model to capitalize on the preferred directions to amplify useful token classes.
We break the V-O and Q-K symmetries by adding random, non-learned, per-batch query and value biases.  
Within the attention head (\ref{eq-attention}) we replace
\begin{equation}\label{eq-v-and-q-with-bs}
    \begin{aligned}
    v &= \WV x \to \WV x + \bV(\text{batch}),\\
    q &= W_Q x\to W_Qx + \bQ(\text{batch}),
    \end{aligned}
\end{equation}
where the components of $\bV$ and $\bQ$ are  are resampled independently for each training batch, from normal distributions $\mathcal{N}(\mu_V, \sigma_V^2 )$ and $\mathcal{N}(\mu_Q, \sigma_Q^2 )$ respectively. 
The nonzero variance in the batchwise biaes enable us to fully break the symmetry during the course of training: to break an $O(d)$ rotational  symmetry, we must introduce at least $d$ independent bias directions; a single prefered direction alone breaks $O(d)$ only down to $O(d-1)$.  Key biases cancel out of the attention (\ref{eq-attention}), and $b_V$ suffices to break the V-O symmetry.  
During inference, the mean biases are applied.  Algorithm \ref{alg:symmetry-broken-attention} shows the forward pass. 

\begin{algorithm}[tb]
\caption{Full Symmetry-Broken Attention}
\label{alg:symmetry-broken-attention}
\begin{algorithmic}[1]
    \STATE \textbf{Input:} $x \in \mathbb{R}^{B \times T \times C}$
    \IF{training \AND mode = ``per\_batch''}
        \STATE Resample $\bQ \sim \mathcal{N}(\mu_Q, \sigma_Q^2)$ \COMMENT{Q-K breaking}
        \STATE Resample $\bV \sim \mathcal{N}(\mu_V, \sigma_V^2)$ \COMMENT{V-O breaking}
    \ELSE
        \STATE $\bQ \gets \mu_Q$, \quad $\bV \gets \mu_V$ \COMMENT{Use means at inference}
    \ENDIF
    \STATE $q, k, v \gets \text{split}(\WQ x, \WK x, \WV x)$
    \STATE $q \gets q + \bQ$, $v \gets v + \bV$  \COMMENT{Add untrained biases}
    \STATE $\text{scores} \gets (q \cdot k^\top) / \sqrt{d_k}$
    \STATE $\text{att} \gets \text{softmax}(\text{scores} + \text{causal\_mask})$
    \STATE \textbf{return} $\WO(\text{att} \cdot v)$
\end{algorithmic}
\end{algorithm}
  
Specifically,  we take nonzero $b_Q$ mean $\mu_Q=0.5$ and consider both choices $\mu_V=0, 0.5$ for the value bias.  The standard deviation in the case of $b_Q$ is taken to linearly increase from $0.05$ to $0.15$ across the 64 dimensions of each head; the fixed std for $b_V$ is taken from a scan of two values $[0.02, 0.05]$ for each optimizer.  
These choices can all be varied, with a systematic ablation study deferred to followup work. 
The current experiments suffice to show strong effects of this particular symmetry breaking protocol.  

As we will spell out in \S\ref{sec:bq_alignment}, this protocol introduces an interpretability benefit of the factor of $e^{ b_Q^T W_Kx_j}$ now multiplying attention (\ref{eq-attention}).  With the nonzero mean $ \mathbb{E} [b_Q] = (0.5,\dots, 0.5)$ introducing a preferred direction, the model can tune $W_K$ such that for particular classes of embedded tokens $x_*$,  $W_K x_*$ align or anti-align with $\mathbb{E}[b_Q]$.




\section{Empirical Evaluation and interpretability}
\label{sec:empirical}

We evaluate our symmetry-breaking protocol by pretraining modified GPT-2 (124M) models on FineWeb-Edu for 500M tokens, comparing four optimizers---ECD, SGDM\footnote{Nesterov=True in the Pytorch implementation.}, AdamW, and SOAP---under symmetric, $b_Q$-only, and full $b_Q + b_V$ symmetry breaking configurations with unlearned batchwise biases.\footnote{In appendix \ref{app:learned_biases} we study models trained with learned biases, mainly initialized with nonzero mean. This restores the symmetry since the learned biases can rotate rather than introducing a preferred direction.}  For all models (including those denoted symmetric) we implement either PreLU \cite{He_2015_ICCV} activations in the MLP block which allow for learned activation asymmetries, or GELU activations.  We vary only the degree of symmetry breaking in the attention blocks.  We assess both validation loss and downstream reasoning via a suite of 14 logic puzzle tasks spanning pattern completion, in-context retrieval, and simple inference. We test whether the models learn to use the preferred mean $b_Q$ direction to amplify or suppress identifiable token classes, and use this to elucidate the performance results.   


Our sample sizes are small due to resource limitations; these results illustrate the effect without extensive statistics or analysis of scaling with model size, data or compute.

\subsection{Symmetry Breaking Enables Efficient Optimization}
\label{sec:performance}

Tables~\ref{tab:main_compressed} and \ref{tab:ecd_seeds}  illustrate that in the modified GPT2 models just defined, our symmetry breaking method by unlearned biases in the attention heads renders ECD competitive with adaptive optimizers.  Among two seeds (42, 123) in the symmetric case, ECD's best validation loss (3.93) is substantially worse than (3.38/4.49) for Adam/SOAP. With full $b_Q + b_V$ symmetry breaking (here with $\mu_V=0$), ECD improves to (3.35)--- nearly matching SOAP (3.33) and substantially outperforming its symmetric baseline.   
SOAP, SGDM, and ECD improve under symmetry breaking, fitting with the fact that they are the optimizers which at least approximately respect the rotational symmetry, whereas AdamW does not.  


\begin{table}[h]
\centering
\caption{\textbf{Illustration of symmetry-breaking improvement} Best validation loss by optimizer and symmetry condition (500M tokens, GPT-124M), among 2 seeds for each optimizer using hyper-parameters given in Appendix \ref{app-HPs},  here considering zero mean for $b_V, \mu_V=0$. $\Delta$ columns show improvement from symmetric baseline (negative = better). Bold indicates best result per optimizer among this small sample. SGDM is run with nonzero initial velocity in order to probe its response to breaking the symmetry conserved currents as in \S\ref{sec-ECD-SGDM-conserved-angular-momenta}. (SGDM with zero initial velocity reaches 3.85 in symmetric mode, nearly identical to the nonzero-velocity baseline).}
\label{tab:main_compressed}
\begin{tabular}{lccc}
\toprule
Optimizer & Symmetric & $b_Q + b_V$ & $\Delta$ \\
\midrule
ECD   & 3.93 & \textbf{3.35} & $-0.58$ \\
SGDM  & 3.84 & \textbf{3.67} & $-0.17$ \\
AdamW & \textbf{3.38} & 3.53 & $+0.15$ \\
SOAP  & 3.49 & \textbf{3.33} & $-0.16$ \\
\bottomrule
\end{tabular}
\end{table}

This improvement for ECD also arises with nonzero $b_V$ mean  $\mu_V = 0.5$, for which it is robust across several random seeds as in Table~\ref{tab:ecd_seeds}.  We keep a record of each seed's results, as we will later analyze their downstream logic capabilities and $b_Q$-key vector alignment properties individually.


\begin{table}[h]
\centering
\caption{ECD validation loss across four seeds, comparing symmetric to $b_Q + b_V$, in this case with $\mu_V = 0.5$.}
\label{tab:ecd_seeds}
\begin{tabular}{lccc}
\toprule
Seed & Symmetric & $b_Q + b_V$ & $\Delta$ \\
\midrule
42  & 3.93 & 3.37 & $-0.56$ \\
83  & 3.82 & 3.40 & $-0.42$ \\
123 & 3.94 & 3.37 & $-0.57$ \\
789 & 3.81 & 3.40 & $-0.41$ \\
\midrule
Mean & 3.88 & 3.39 & $-0.49$ \\
Std  & 0.07 & 0.02 & $0.09$ \\
\bottomrule
\end{tabular}
\end{table}

\textbf{Changing the MLP block} activation function and symmetry structure leads to smaller improvement.
We conduct an ablation study replacing PReLU -- a choice with trainable activation slopes which enables the model to break the MLP block permutation symmetry --  with the GELU activation, a more common choice for the GPT2 architecture. We train using ECD with three seeds (42, 83, 789) across symmetric and symmetry-broken configurations, plus additional optimizer comparisons at seed 42. GELU models achieve substantially lower validation loss than PReLU across all configurations (3.17 vs 3.88 for symmetric baselines). 
However, symmetry breaking provides a much smaller improvement on the overall validation loss for GELU: $\Delta = -0.03$ (std $0.018$) compared to $\Delta = -0.49$ (std $0.09$)  for PReLU (Table~\ref{tab:gelu_val_loss}). 

\textbf{Examining logic puzzle performance.}
While validation loss improvements are consistent, effects on downstream reasoning are more heterogeneous. Table~\ref{tab:combined_logic} summarizes logic puzzle performance (Top-5 accuracy) for all ECD models. This includes the same four ECD seeds with the PreLU MLP activation as in Table \ref{tab:ecd_seeds}. Two seeds (42, 83) show improvement with symmetry breaking, one (123) degrades, and one (789) is unchanged.  Similarly for the GELU MLP case, we find 5/8 tested seeds show improvement with symmetry breaking, 2/8 are neutral and 1 is harmful (see Table \ref{tab:gelu_logic}.  Overall, we have a mean improvement on top 5 accuracy of $4.7 \pm 8.79$.  

This is not a statistically significant improvement on the downstream task.  Still, the majority of models improve on logic, and we would like to understand what distinguishes those that benefit from the introduction of the $b_Q, b_V$ biases from those that are neutral or harmed by them (only $16.7\%$ overall).  Since the validation loss improvements are robust, that is not a predictor of logic puzzle improvement at our model and data scale.  Interestingly, we will find that the alignment mechanism we anticipated theoretically in \S\ref{sec-rotation-breaking-bQ-bV} provides a better predictor of logic puzzle success than does the overall best validation loss.\footnote{In Appendix \ref{app:logic_puzzles} we include more details on the protocol and results including comparison with other optimizers.
}   

\begin{table}[h]
\centering
\caption{Combined logic puzzle performance for PReLU (P) with $\mu_V=0.5$ and GELU (G) ECD models with a variety of symmetry breaking settings.}
\label{tab:combined_logic}
\small
\begin{tabular}{cllcc}
\toprule
\textbf{Act.} & \textbf{Seed} & \textbf{Config} & \textbf{SB Top-5} & \textbf{$\Delta$ Top-5} \\
\midrule
P & 42  & $b_Q{+}b_V$  & 70.8\% & $+4.1$ \\
P & 83  & $b_Q{+}b_V$  & 66.7\% & $+4.2$ \\
P & 123 & $b_Q{+}b_V$  & 62.5\% & $-8.3$ \\
P & 789 & $b_Q{+}b_V$  & 62.5\% & $+0.0$ \\
\midrule
G & 42  & bQonly       & 71.4\% & $+0.0$ \\
G & 42  & bQbV-V00     & 78.6\% & $+7.1$ \\
G & 83  & bQonly       & 78.6\% & $+0.0$ \\
G & 83  & bQbV-V00     & 85.7\% & $+7.1$ \\
G & 83  & bQbV-V05     & 71.4\% & $-7.1$ \\
G & 789 & bQonly       & 78.6\% & $+14.3$ \\
G & 789 & bQbV-V00     & 85.7\% & $+21.4$ \\
G & 789 & bQbV-V05     & 78.6\% & $+14.3$ \\
\midrule
\multicolumn{4}{l}{Mean} & $+4.76$ \\
\multicolumn{4}{l}{Std}  & $8.79$ \\
\bottomrule
\end{tabular}
\end{table}



\subsection{Interpretable $b_Q$ Alignment}
\label{sec:bq_alignment}
The symmetry-breaking mechanism provides more than an optimization benefit at the kinematical level. In addition, the fixed direction $\mathbb{E}[b_Q]$ becomes an preferred axis that models can learn to exploit. As described in \S\ref{sec-rotation-breaking-bQ-bV}, the query bias enters the attention computation as a multiplicative factor $e^{k \cdot b_Q}$, enabling the model to systematically amplify or suppress attention to tokens based on the alignment between their key vectors $k = W_K x$ and $\mathbb{E}[b_Q]$.  
In this section, we will analyze this and use it to help understand variations in logic puzzle performance. 

\textbf{Protocol.}
For each trained model, we compute the cosine similarity between the mean query bias $\mathbb{E}[b_Q]$ and the key projection $W_K x$ for all tokens $x$ in the vocabulary:
\begin{equation}
    \text{alignment}(x) = \frac{(W_K x)^\top \mathbb{E}[b_Q]}{\|W_K x\| \, \|\mathbb{E}[b_Q]\|}
\end{equation}
For each of the 144 attention heads (12 layers $\times$ 12 heads), we identify the 15 tokens with highest alignment (``top-aligned,'' attention-enhanced) and lowest alignment (``bottom-aligned,'' attention-suppressed), yielding 2,160 tokens in each category per model.  The input to a given attention layer is the full residual stream:
   $ x_L = x_{tok} + x_{pos} + \sum_{\ell=0}^{L-1} (\text{Attn}_\ell + \text{MLP}_\ell)$.
For all layers there is a factor $\exp(b_Q W_K x)$.  

\textbf{Alignment far exceeds chance.}
Under a null hypothesis of random orientation in $d = 64$ dimensions with vocabulary size $V = 50{,}304$, extreme value theory predicts a maximum alignment of approximately 0.58, and empirically
ECD Symmetric models have a mean top alignment 0.488 and maximum top alignment of 0.572, fitting with this null baseline.  ECD models with $b_Q + b_V$ achieve mean top alignment of 0.73, with 97\% of attention heads exceeding the random threshold.  This confirms that models actively learn to align $W_K$ with the fixed $\mathbb{E}[b_Q]$ direction. 

\textbf{Semantic enrichment.}
Inspection of the top aligned tokens reveals striking patterns; Table \ref{tab:example_head} illustrates this with an anectodal example.  We now perform a systematic analysis of the effect and its relation to the heterogeneous logic puzzle results.    
\begin{table}[h]
\centering
\caption{Tokens with highest/lowest alignment with $\mathbb{E}[\bQ]$ in an ECD-trained model with K-Q symmetry breaking trained on 500M tokens for a Layer 0 attention head. The model learns to enhance attention to sentence-initial tokens that signal logical structure.}
\label{tab:example_head}
\begin{tabular}{lr|lr}
\toprule
\multicolumn{2}{c|}{\textbf{Top Aligned}} & \multicolumn{2}{c}{\textbf{Bottom Aligned}} \\
\textbf{Token} & \textbf{Sim.} & \textbf{Token} & \textbf{Sim.} \\
\midrule
\texttt{Considering} & $+0.936$ & \texttt{oll} & $-0.881$ \\
\texttt{Given} & $+0.936$ & \texttt{oj} & $-0.862$ \\
\texttt{Born} & $+0.935$ & \texttt{rim} & $-0.861$ \\
\texttt{Overall} & $+0.931$ & \texttt{arn} & $-0.861$ \\
\texttt{Disclaimer} & $+0.926$ & \texttt{alk} & $-0.861$ \\
\texttt{Currently} & $+0.925$ & \texttt{mot} & $-0.857$ \\
\texttt{Assuming} & $+0.923$ & \texttt{ood} & $-0.857$ \\
\texttt{Nonetheless} & $+0.920$ & \texttt{eal} & $-0.855$ \\
\texttt{Interestingly} & $+0.920$ & \texttt{oles} & $-0.853$ \\
\texttt{Posted} & $+0.920$ & \texttt{ux} & $-0.849$ \\
\bottomrule
\end{tabular}
\end{table}
To characterize \emph{what} models learn to enhance or suppress, we classify tokens into semantic categories and compute the log$_2$ enrichment ratio:
\begin{equation}\label{eq:enrichment-ratio}
    \text{enrichment} = \log_2 \frac{f_{\text{top}}}{f_{\text{bottom}}}
\end{equation}
where $f_{\text{top}}$ and $f_{\text{bottom}}$ are the frequencies of a category among top-aligned and bottom-aligned tokens. Positive values indicate that a category is enriched among attention-enhanced tokens; negative values indicate attention-suppression.  


All four ECD PreLU models learn consistent semantic patterns (Table~\ref{tab:semantic_revised}): enhancing structural markers (sentence starters, interrogatives, punctuation, function words) while suppressing unicode artifacts and encoding errors, and similarly for GELU (Tables \ref{tab:semantic-gelu}, \ref{tab:punct_example}).  Focusing on our 12 listed PreLU and GELU ECD models, three  nearly universal enrichment patterns reach statistical significance overall (binomial test): sentence starters and interrogatives enhanced (12/12, p < 0.001; 11/12, p = 0.003), noise suppressed (11/12, p = 0.003).  

\textbf{Alignment patterns and reasoning.}
The variable effect of symmetry breaking on logic puzzles (Table~\ref{tab:combined_logic}) correlates with specific alignment characteristics, elucidating the heterogeneous logic results in \S\ref{sec:performance}.  Let us first examine this for the PreLU MLP models. Seeds that benefit (42, 83) exhibit stronger noise suppression 
compared to the hurt and neutral seeds,
and moderate-to-strong punctuation enrichment. 
Seed 123, which is hurt by symmetry breaking, shows drastically weaker punctuation enrichment 
and, at Layer 0, extreme function word suppression. 
This suggests that effective $b_Q$ alignment requires  enhancing attention to structural markers as well as cleaning attention away from noise. 
Across GELU models, we see a similar pattern: punctuation enrichment correlates strongly with logic puzzle benefit (Table~\ref{tab:gelu_punct}).\footnote{SOAP benefits despite weak alignment ($0.28$) because its alignment enhances structural tokens (enrichment $+0.14$), while Adam is hurt despite strong alignment ($0.66$) because it suppresses them (enrichment $-1.70$).} : Pearson $r = 0.76$ for ECD ($n=8$, 95$\%$ confidence interval [+0.12, +0.95]) and $r = 0.72$ across all optimizers ($n=11$, 95$\%$ confidence interval [+0.21, +0.92]). Models that benefit from symmetry breaking show positive punctuation enrichment, 
while hurt models show negative enrichment 
Altogether, our results fit with the idea that symmetry breaking helps reasoning when the model learns to align useful tokens' key vectors with $\mathbb{E}[ \mathbf{b}_Q]$.



\begin{table}[t]
\caption{Alignment quality predicts logic puzzle benefit. For each model (including all optimizers), we compute punctuation enrichment as the log$_2$ ratio of structural token frequency in top- vs bottom-aligned vocabulary, and report mean alignment and punctuation enrichment. Pearson correlation is computed across models between enrichment and $\Delta$ Top-5.}
\label{tab:gelu_punct}
\centering
\small
\begin{tabular}{lccc}
\toprule
\textbf{Effect} & \textbf{$n$} & \textbf{Align} & \textbf{Punct Enrich} \\
\midrule
Benefit & 6 & 0.63 & $+0.34$ \\
Neutral & 2 & 0.70 & $-0.99$ \\
Hurt    & 3 & 0.61 & $-0.91$ \\
\midrule
\multicolumn{2}{l}{Pearson $r$ with $\Delta$ Top-5:} & $0.23$ & $0.72$ \\
\bottomrule
\end{tabular}
\end{table}




\subsection{Summary of empirical evaluation}

Symmetry breaking consistently improves ECD's validation loss across all seeds -- especially with the GPT2 124M prelu-MLP model --allowing the memory-efficient Hamiltonian optimizer to match adaptive methods, with smaller gains observed for the GeLU MLP.  Logic puzzle performance improves or remains unchanged in a strong majority of seeds, with the variable outcomes predictable from semantic $b_Q$ alignment patterns.  The $b_Q$ mechanism provides an interpretable window into how models exploit the symmetry breaking, learning to enhance attention to discourse markers or punctuation and suppress noise.





\section{Discussion and Limitations}

In this work we demonstrated how introducing a theoretically motivated  $b_Q, b_V$ symmetry breaking protocol into transformer attention leads to efficient and interpretable training.  We investigated its effects on a suite of optimizers, with ECD (a predictive, memory-saving option) particularly benefiting from symmetry breaking, reaching competitive performance with more complex methods.  
The empirical results in this work are limited by having small statistics and modest parameter count, data, and compute.   In the future work it will be important to scale these up, with additional ablations. Additional theory would also be useful to help determine in a more principled way the optimal means and variances of the biases.  

A very interesting direction for future work concerns the interpretable key-bias alignments and their role in predicting downstream reasoning performance.  We would like to understand this as a function of layer, exploring alignment of $b_Q$ with the full residual stream.

\section*{Acknowledgements}



We would like to thank G. B. De Luca, Alice Gatti, B. Nachman and H. Zheng many useful discussions and recent collaboration on ECD.  We also thank David Africa, Ven Chandrasekaran, Mehmet Demirtas, Ethan Dyer, Surya Ganguli, Indranil Halder, Jim Halverson, Dan Roberts, and Jakob Robnik for very insightful comments.  As detailed in App. \ref{app:LLM-use}, Claude code was instrumental in carrying out this research and we thank Anthropic for a trial subscription.  
The research of ES was supported by a Simons Investigator Award and by the Simons Collaboration on the Physics of Learning and Neural Computation.
The work of DK was partially supported by the Miller Institute for Basic Research in Science, University of California, Berkeley.





\begin{thebibliography}{27}
\providecommand{\natexlab}[1]{#1}
\providecommand{\url}[1]{\texttt{#1}}
\expandafter\ifx\csname urlstyle\endcsname\relax
  \providecommand{\doi}[1]{doi: #1}\else
  \providecommand{\doi}{doi: \begingroup \urlstyle{rm}\Url}\fi

\bibitem[Bayer et~al.(2023)Bayer, Seljak, and Modi]{BayerSeljakModi2023}
Bayer, A.~E., Seljak, U., and Modi, C.
\newblock Field-level inference with {M}icrocanonical {L}angevin {M}onte
  {C}arlo.
\newblock In \emph{ICML Workshop on Machine Learning for Astrophysics}, 2023.
\newblock URL \url{https://arxiv.org/abs/2307.09504}.

\bibitem[De~Luca \& Silverstein(2022)De~Luca and Silverstein]{BBI}
De~Luca, G.~B. and Silverstein, E.
\newblock Born-infeld ({BI}) for {AI}: Energy-conserving descent ({ECD}) for
  optimization.
\newblock In Chaudhuri, K., Jegelka, S., Song, L., Szepesvari, C., Niu, G., and
  Sabato, S. (eds.), \emph{Proceedings of the 39th International Conference on
  Machine Learning}, volume 162 of \emph{Proceedings of Machine Learning
  Research}, pp.\  4918--4936. PMLR, 17--23 Jul 2022.
\newblock URL \url{https://proceedings.mlr.press/v162/de-luca22a.html}.

\bibitem[De~Luca et~al.(2023)De~Luca, Gatti, and Silverstein]{DeLuca:2023yld}
De~Luca, G.~B., Gatti, A., and Silverstein, E.
\newblock {Improving Energy Conserving Descent for Machine Learning: Theory and
  Practice}.
\newblock 6 2023.

\bibitem[De~Luca et~al.(2025)De~Luca, Nachman, Silverstein, and
  Zheng]{DeLuca:2025ruv}
De~Luca, G.~B., Nachman, B., Silverstein, E., and Zheng, H.
\newblock {Optimizers for stabilizing likelihood-free inference}.
\newblock \emph{Phys. Rev. D}, 112\penalty0 (9):\penalty0 092008, 2025.
\newblock \doi{10.1103/x88j-mv39}.

\bibitem[Dirac(1964)]{Dirac1964}
Dirac, P. A.~M.
\newblock \emph{Lectures on Quantum Mechanics}.
\newblock Number~2 in Belfer Graduate School of Science Monographs Series.
  Belfer Graduate School of Science, Yeshiva University, New York, 1964.

\bibitem[Gupta et~al.(2018)Gupta, Koren, and Singer]{Guptaetal2018}
Gupta, V., Koren, T., and Singer, Y.
\newblock Shampoo: Preconditioned stochastic tensor optimization.
\newblock In \emph{International Conference on Machine Learning (ICML)}, pp.\
  1842--1850, 2018.

\bibitem[He et~al.(2015)He, Zhang, Ren, and Sun]{He_2015_ICCV}
He, K., Zhang, X., Ren, S., and Sun, J.
\newblock Delving deep into rectifiers: Surpassing human-level performance on
  imagenet classification.
\newblock In \emph{Proceedings of the IEEE International Conference on Computer
  Vision (ICCV)}, pp.\  1026--1034, 2015.
\newblock \doi{10.1109/ICCV.2015.123}.

\bibitem[Karpathy(2022)]{karpathy2022nanogpt}
Karpathy, A.
\newblock nanogpt: The simplest, fastest repository for training/finetuning
  medium-sized gpts, 2022.
\newblock URL \url{https://github.com/karpathy/nanoGPT}.
\newblock Accessed: 2026-01-25.

\bibitem[Kingma \& Ba(2015)Kingma and Ba]{KingmaBa2015}
Kingma, D.~P. and Ba, J.
\newblock Adam: A method for stochastic optimization.
\newblock In \emph{International Conference on Learning Representations
  (ICLR)}, 2015.

\bibitem[{Kunin} et~al.(2020){Kunin}, {Sagastuy-Brena}, {Ganguli}, {Yamins},
  and {Tanaka}]{2020arXiv201204728K}
{Kunin}, D., {Sagastuy-Brena}, J., {Ganguli}, S., {Yamins}, D. L.~K., and
  {Tanaka}, H.
\newblock {Neural Mechanics: Symmetry and Broken Conservation Laws in Deep
  Learning Dynamics}.
\newblock \emph{arXiv e-prints}, art. arXiv:2012.04728, December 2020.
\newblock \doi{10.48550/arXiv.2012.04728}.

\bibitem[Liu et~al.(2024)Liu, Li, Hall, Liang, and Ma]{Liuetal2024Sophia}
Liu, H., Li, Z., Hall, D., Liang, P., and Ma, T.
\newblock Sophia: A scalable stochastic second-order optimizer for language
  model pre-training.
\newblock In \emph{International Conference on Learning Representations
  (ICLR)}, 2024.
\newblock URL \url{https://openreview.net/forum?id=3xHDeA8Noi}.

\bibitem[Liu et~al.(2025)Liu, Su, Yao, Jiang, Lai, Du, Qin, Xu, Lu, Yan,
  et~al.]{Liuetal2025}
Liu, J., Su, J., Yao, X., Jiang, Z., Lai, G., Du, Y., Qin, Y., Xu, W., Lu, E.,
  Yan, J., et~al.
\newblock Muon is scalable for llm training.
\newblock \emph{arXiv preprint arXiv:2502.16982}, 2025.

\bibitem[Loshchilov \& Hutter(2019)Loshchilov and Hutter]{LoshchilovHutter2019}
Loshchilov, I. and Hutter, F.
\newblock Decoupled weight decay regularization.
\newblock In \emph{International Conference on Learning Representations
  (ICLR)}, 2019.

\bibitem[Noether(1918)]{Noether1918}
Noether, E.
\newblock Invariante variationsprobleme.
\newblock \emph{Nachrichten von der Gesellschaft der Wissenschaften zu
  G\"ottingen, Mathematisch-Physikalische Klasse}, 1918:\penalty0 235--257,
  1918.

\bibitem[Polyak(1964)]{Polyak1964}
Polyak, B.~T.
\newblock Some methods of speeding up the convergence of iteration methods.
\newblock \emph{USSR Computational Mathematics and Mathematical Physics},
  4\penalty0 (5):\penalty0 1--17, 1964.
\newblock \doi{10.1016/0041-5553(64)90137-5}.

\bibitem[Robbins \& Monro(1951)Robbins and Monro]{RobbinsMonro1951}
Robbins, H. and Monro, S.
\newblock A stochastic approximation method.
\newblock \emph{The Annals of Mathematical Statistics}, 22\penalty0
  (3):\penalty0 400--407, 1951.
\newblock \doi{10.1214/aoms/1177729586}.

\bibitem[Robnik et~al.(2023)Robnik, De~Luca, Silverstein, and
  Seljak]{robnik2023microcanonical}
Robnik, J., De~Luca, G.~B., Silverstein, E., and Seljak, U.
\newblock Microcanonical hamiltonian monte carlo.
\newblock \emph{Journal of Machine Learning Research}, 24\penalty0
  (311):\penalty0 1--34, 2023.

\bibitem[Sharkey et~al.(2025)Sharkey, Chughtai, Batson, Lindsey, Wu, Bushnaq,
  Goldowsky-Dill, Heimersheim, Ortega, Bloom, et~al.]{sharkey2025open}
Sharkey, L., Chughtai, B., Batson, J., Lindsey, J., Wu, J., Bushnaq, L.,
  Goldowsky-Dill, N., Heimersheim, S., Ortega, A., Bloom, J., et~al.
\newblock Open problems in mechanistic interpretability.
\newblock \emph{arXiv preprint arXiv:2501.16496}, 2025.

\bibitem[Shazeer \& Stern(2018)Shazeer and Stern]{ShazeerStern2018}
Shazeer, N. and Stern, M.
\newblock Adafactor: Adaptive learning rates with sublinear memory cost.
\newblock In \emph{International Conference on Machine Learning (ICML)}, pp.\
  4596--4604, 2018.

\bibitem[Sommer et~al.(2025)Sommer, Robnik, Nozadze, Seljak, and
  R{\"u}gamer]{Sommeretal2025}
Sommer, E., Robnik, J., Nozadze, G., Seljak, U., and R{\"u}gamer, D.
\newblock Microcanonical {L}angevin ensembles: Advancing the sampling of
  {B}ayesian neural networks.
\newblock In \emph{International Conference on Learning Representations
  (ICLR)}, 2025.

\bibitem[Susskind \& Hrabovsky(2013)Susskind and
  Hrabovsky]{SusskindHrabovsky2013}
Susskind, L. and Hrabovsky, G.
\newblock \emph{The Theoretical Minimum: What You Need to Know to Start Doing
  Physics}.
\newblock Basic Books, New York, 2013.
\newblock ISBN 978-0465028115.

\bibitem[Tanaka \& Kunin(2021)Tanaka and Kunin]{tanaka2021noether}
Tanaka, H. and Kunin, D.
\newblock Noether’s learning dynamics: Role of symmetry breaking in neural
  networks.
\newblock \emph{Advances in Neural Information Processing Systems},
  34:\penalty0 25646--25660, 2021.

\bibitem[Tieleman \& Hinton(2012)Tieleman and Hinton]{TielemanHinton2012}
Tieleman, T. and Hinton, G.
\newblock Lecture 6.5---rmsprop: Divide the gradient by a running average of
  its recent magnitude.
\newblock COURSERA: Neural Networks for Machine Learning, 2012.

\bibitem[Vyas et~al.(2024)Vyas, Jabbour, Lajoie, and Bengio]{Vyasetal2024}
Vyas, N., Jabbour, A., Lajoie, G., and Bengio, Y.
\newblock Soap: Improving and stabilizing shampoo using adam for language
  modeling.
\newblock In \emph{Advances in Neural Information Processing Systems
  (NeurIPS)}, 2024.

\bibitem[Wen et~al.(2025)Wen, Hall, Ma, and Liang]{wen2025fantastic}
Wen, K., Hall, D., Ma, T., and Liang, P.
\newblock Fantastic pretraining optimizers and where to find them.
\newblock \emph{arXiv preprint arXiv:2509.02046}, 2025.

\bibitem[Wibisono et~al.(2016)Wibisono, Wilson, and
  Jordan]{wibisono2016variational}
Wibisono, A., Wilson, A.~C., and Jordan, M.~I.
\newblock A variational perspective on accelerated methods in optimization.
\newblock \emph{proceedings of the National Academy of Sciences}, 113\penalty0
  (47):\penalty0 E7351--E7358, 2016.

\bibitem[Zhang et~al.(2025)Zhang, Zheng, Chen, and Li]{zhang2025beyond}
Zhang, B., Zheng, Z., Chen, Z., and Li, J.
\newblock Beyond the permutation symmetry of transformers: The role of rotation
  for model fusion.
\newblock \emph{arXiv preprint arXiv:2502.00264}, 2025.

\end{thebibliography}
\bibliographystyle{icml2026}

\newpage
\appendix
\onecolumn

\section{Use of LLMs}\label{app:LLM-use}

This work included significant use of Claude Code to write much of the code package implementing the empirical research plan. The theory and the empirical research plan itself was author-generated. Claude code came up with the particular set of logic puzzles we use as a downstream reasoning test, as well as proposing the particular set of predefined semantic categories we use for the alignment analysis, prompted by the authors' detailed instructions to set up such logic and alignment analyses. This code, including model construction and training building from and modifying the nano-GPT2 constructions by \cite{karpathy2022nanogpt} and  in the modded-nanogpt repo \href{https://github.com/KellerJordan/modded-nanogpt}{here},   was tested, cross-checked, and corrected in various ways by the human authors.   Claude was also used to assemble and statistically analyze results on alignment and logic puzzle performance, and also to generate LaTeX code for tables and several text subsections (then heavily edited by the authors). 

\section{The $ECD_{q=1}$ version of ECD optimization}
\label{app:ECD-variants}

ECD optimization is a general framework with a variety of realizations \cite{BBI, DeLuca:2023yld, DeLuca:2025ruv}.  In general,  its update rules are a discretization of energy-conserving Hamiltonian dynamics, with Hamiltonian chosen so that the process concentrates results (converges at) low loss values.  Examples include the variable mass Hamiltonian discussed above
\begin{equation}\label{eq-ECD-variable-mass-Ham}
  H=  \frac{\Pi^2}{2m} = \frac{1}{2} m(\Theta) \dot\Theta^2 = E
\end{equation}
with mass inversely related to the loss $F(\Theta)$, e.g. $m(\Theta) = (F(\Theta)-F_0)^{-\eta}$.  A different version fitting into this framwork, which performs effectively in precicion science applications,  is $ECD_{q=1}$ \cite{DeLuca:2025ruv, robnik2023microcanonical} with Hamiltonian,
\begin{equation}\label{eq-ECDq=1-Hamiltonian}
    H = |\Pi| -(F({\bf{\Theta}})-F_0)^{-\frac{d\eta}{2(d-1)}} + E
\end{equation}
In both cases, the Liouville measure which captures the distribution on phase space, when integrated over momentum, gives a distribution on parameter space 
\begin{align}\nonumber\label{eq-measure}
    p({\bf{\Theta}}) &\propto \int_{\mathbb{R}^d}\delta(H({\bf{\Theta}},{\bf{\Pi}})-E)|\Pi|^{d-1}d\Pi \\
    &\hspace{8mm}= (F({\bf{\Theta}}) - F_0)^{-\frac{\eta d}{2}}\, .
\end{align}
which concentrates toward lower loss increasingly with increasing $\eta$.  
Limitations on how large a value of $\eta$ is useful as a function of initialization and compute budget were motivated in \cite{DeLuca:2025ruv} for the $ECD_{q=1}$ case.  

 There is a nuance that this particular Hamiltonian (\ref{eq-ECDq=1-Hamiltonian}) describes a constrained system since it has the property that $\partial H/\partial\Pi\partial\Pi$ is not invertible \cite{Dirac1964}.  Let us spell this out, which requires introducing the Lagrangian (a relative of the Hamiltonian) in order to understand the relation between momenta $\Pi$ and velocities $\dot\Theta$. 

The Lagrangian $L(q, \dot q, t)$ is a function of positions and velocities (and in the case without energy conservation, it would also be an explicit function of time).  From $L$ one can derive the equations of motion by a variational principle, enforcing that the variation of $\int dt L$ with respect to $q(t)$ vanish.\footnote{For a pedagogical introduction see e.g. \cite{SusskindHrabovsky2013}.}  The momenta are defined by varying $L$ with respect to $\dot q$:  $\Pi_q = \partial L/\partial \dot q$, and the Hamiltonian is obtained from the Lagrangian as $H = \sum_I \Pi_{q_I} \dot q_I -L(t)$, substituting $\dot q$ for its expression in terms of $\Pi_q$.

We will spell this out directly in our case of interest.  We start from a Lagrangian 
\begin{equation}\label{eq-q=1-Lagrangian}
    L=\frac{1}{2}\lambda(\dot\Theta^2-1) + (F(\Theta)-F_0)^{-d\eta/(d-1)}-E
\end{equation}
which in addition to $\Theta$ has an additional variable $\lambda$ (known as a Lagrange multiplier) which enforces constant velocity: 
\begin{equation}\label{eq-q=1-const-velocity-constraint}
\delta_\lambda \int dt L \equiv 0 \Rightarrow \dot\Theta^2=\dot W_Q^T\dot W_Q + \dot W_K^T \dot W_K + \dot W_V^T\dot W_V + \dot W_O^T \dot W_O+\dot\Theta_\perp^2=1.  
\end{equation}
The variable $\lambda$ has zero conjugate momentum, $\Pi_\lambda = \partial L/\partial{\dot\lambda} = 0$.
The conjugate momentum to $\Theta$,  $\Pi$ is given by
\begin{equation}\label{eq-q=1-Pi-froml}
\Pi = \frac{\partial L}{\partial{\dot\Theta}}=\lambda\dot\Theta
\end{equation}
The Hamiltonian is then given by
\begin{eqnarray}
    \label{eq-q=1-Ham-from-Lag}
H &= \sum_I p_I \dot q_I -L = \Pi_\lambda\dot\lambda +\Pi \dot\Theta -L \nonumber \\
&= \frac{\Pi^2}{2\lambda} +\frac{\lambda}{2}-(F(\Theta)-F_0)^{-d\eta/(d-1)} + E
\end{eqnarray}
where $p_I$ and $q_I$ denote momenta and position variables.  From this, applying the Hamilton equations (\ref{eq:hamiltons-eqs}) to this system gives $\dot\Theta=\Pi/\lambda$. Then using (\ref{eq-q=1-const-velocity-constraint}) leads to
\begin{equation}\label{eq-lambda-is-pi}
    \lambda = |\Pi|;
\end{equation}
substituting this into (\ref{eq-q=1-Ham-from-Lag}) we recover the ECD$_{q=1}$ Hamiltonian (\ref{eq-ECDq=1-Hamiltonian}). 

\section{Rotational Drift Analysis}
\label{app:rotational_drift}

The kinematic effect of the symmetry breaking is dependent on there being nontrivial angular momentum.  In principle, at $\nu=0$, this is avoided by ECD's initialization along the direction of the gradient.  But numerical errors can build up, generating some angular motion.  In this appendix we describe our measurements of this.  For an interesting precedent for this, see Figure 8 in \cite{2020arXiv201204728K}.

To quantify this effect, we define and measure a {rotational velocity fraction} $f_{rot}$ in our ECD-trained models, the fraction of optimization velocity in rotational directions versus meaningful directions that change the rotation-invariant quantities $G_{QK} = W_Q W_K^T$ and $G_{VO} = W_V W_O$. We analyze both the Q-K sector (with $O(d_k)$ symmetry) and the V-O sector (with $O(d_v)$ symmetry). 


Our measurements reveal some drift in angular momentum, with a notable asymmetry between sectors. In symmetric (no bias) models, the Q-K sector shows minimal rotational velocity ($f_{rot}^{QK} \approx 0.1\%$--$0.4\%$), while the V-O sector shows substantial rotational motion ($f_{rot}^{VO} \approx 11\%$--$14\%$). The total angular momentum magnitudes reflect this: $|J_{VO}|/|J_{QK}| \approx 24$--$33$ in symmetric models. Both sectors show significant drift in $|J|$ over training ($+190\%$ for Q-K, $+114\%$ for V-O from 100M to 500M tokens), confirming that even from near-zero initialization, angular momentum accumulates and can affect optimization.

This hierarchy of effects lines up with an empirical result we found that breaking V-O symmetry (via $b_V$) provides larger improvements than breaking Q-K symmetry (via $b_Q$). 
This correlates with the $\sim 3\times$ larger $f_{rot}$ in the V-O sector. Breaking symmetry converts this ``wasted'' rotational velocity into productive optimization.

This appendix presents detailed measurements of rotational dynamics during ECD training, comparing symmetric models (standard attention) to models with full symmetry breaking ($b_Q$, $b_K$, $b_V$ biases). All models use GPT-124M architecture trained on FineWeb-Edu for 500M tokens with ECD optimizer ($\eta = 100$, $F_0 = 2$).

\subsection{Methodology}

\textbf{Velocity-based angular momentum.} For the Q-K sector with $O(d_k)$ rotational symmetry $W_Q \to W_Q R$, $W_K \to W_K R$, we compute the tensor
\begin{equation}
    L_{QK} = u_Q W_K^T - W_K u_Q^T + u_K W_Q^T - W_Q u_K^T
\end{equation}
where $u = \Pi/|\Pi|$ is the unit velocity (the normalized momentum stored by ECD), and define a rescaled version of the total angular momentum magnitude is $|\tilde{J}_{QK}| = \|L_{QK}\|_F / \sqrt{2}$, summing over all $d_k(d_k-1)/2$ generators of $SO(d_k)$.

This rescaled angular momentum $\tilde{J}$ is related to the true Noether current by $J = |\Pi| \tilde{J}$. Since the ECD prefactor $(F - F_0)^{-d\eta/2(d-1)}$ takes extreme values,
we track $\tilde{J}$ as our diagnostic of rotational motion.

\textbf{Rotational velocity fraction.} We decompose the velocity into rotational and meaningful components. The rotational fraction is
\begin{equation}
    f_{rot}^2 = \frac{d_{head} \cdot |\tilde{J}|^2}{2(\|W_Q\|^2 + \|W_K\|^2)(\|u_Q\|^2 + \|u_K\|^2)}
\end{equation}
where $d_{head} = 64$. This measures what fraction of optimization velocity is ``wasted'' on gauge motion versus changing the gauge-invariant $G = W_Q W_K^T$. The meaningful fraction is $f_{meaningful} = \sqrt{1 - f_{rot}^2}$.

\textbf{V-O sector.} The V-O sector has symmetry $W_V \to W_V \tilde{R}$, $W_O \to \tilde{R}^T W_O$. The angular momentum computation is analogous, accounting for the different index structure of $W_O$.

\subsection{Results}

\textbf{Angular momentum magnitude.} Table~\ref{tab:angular_momentum} shows the total angular momentum in each sector across training checkpoints.

\begin{table}[h]
\centering
\caption{Total angular momentum magnitude $|\tilde{J}|$ by sector and training stage. Symmetric models show much larger V-O angular momentum relative to Q-K.}
\label{tab:angular_momentum}
\begin{tabular}{llccc}
\toprule
\textbf{Model} & \textbf{Sector} & \textbf{100M} & \textbf{250M} & \textbf{500M} \\
\midrule
\multirow{3}{*}{Symmetric} 
& Q-K & 0.0010 & 0.0022 & 0.0030 \\
& V-O & 0.0334 & 0.0526 & 0.0714 \\
& Ratio V-O/Q-K & 32.7 & 23.8 & 24.2 \\
\midrule
\multirow{3}{*}{Symmetry-broken}
& Q-K & 0.0070 & 0.0081 & 0.0096 \\
& V-O & 0.0391 & 0.0600 & 0.0790 \\
& Ratio V-O/Q-K & 5.6 & 7.4 & 8.2 \\
\bottomrule
\end{tabular}
\end{table}

\textbf{Angular momentum drift.} Table~\ref{tab:angular_drift} shows the percentage change in $|\tilde{J}|$ between checkpoints, measuring departure from conservation.

\begin{table}[h]
\centering
\caption{Angular momentum drift (percentage change in $|\tilde{J}|$) over training intervals. Symmetry-broken models show much more stable Q-K angular momentum.}
\label{tab:angular_drift}
\begin{tabular}{llccc}
\toprule
\textbf{Model} & \textbf{Sector} & \textbf{100M$\to$250M} & \textbf{250M$\to$500M} & \textbf{100M$\to$500M} \\
\midrule
\multirow{2}{*}{Symmetric} 
& Q-K & $+117\%$ & $+34\%$ & $+190\%$ \\
& V-O & $+57\%$ & $+36\%$ & $+114\%$ \\
\midrule
\multirow{2}{*}{Symmetry-broken}
& Q-K & $+16\%$ & $+19\%$ & $+38\%$ \\
& V-O & $+54\%$ & $+32\%$ & $+102\%$ \\
\bottomrule
\end{tabular}
\end{table}

\textbf{Rotational velocity fraction.} Table~\ref{tab:velocity_fraction} shows the fraction of velocity in rotational directions.

\begin{table}[h]
\centering
\caption{Rotational velocity fraction $f_{rot}$ by sector and training stage. V-O sector shows substantially higher rotational fraction than Q-K in both model types.}
\label{tab:velocity_fraction}
\begin{tabular}{llccc}
\toprule
\textbf{Model} & \textbf{Sector} & \textbf{100M} & \textbf{250M} & \textbf{500M} \\
\midrule
\multirow{2}{*}{Symmetric} 
& Q-K & 0.13\% & 0.26\% & 0.39\% \\
& V-O & 10.9\% & 12.8\% & 13.9\% \\
\midrule
\multirow{2}{*}{Symmetry-broken}
& Q-K & 5.51\% & 5.02\% & 5.19\% \\
& V-O & 11.8\% & 13.6\% & 14.8\% \\
\bottomrule
\end{tabular}
\end{table}

\textbf{Weight drift ratio.} We also compute the raw weight drift ratio $(||{\Delta W_Q}|| + ||{\Delta W_K}||) / ||{\Delta(W_Q W_K^T)}||$, which measures how much weight changes exceed gauge-invariant changes. Both symmetric and symmetry-broken models show drift ratios $\approx 2$, indicating similar underlying gauge structure regardless of symmetry breaking.

\section{Experimental details}

\subsection{Unified Experimental Setup}

All runs reported in Section~\ref{sec:empirical} use a consistent experimental configuration to enable fair comparisons:

\begin{itemize}
    \item \textbf{Model:} GPT-124M (12 layers, 12 heads, 768 embedding dimension)
    \item \textbf{Vocabulary:} 50,304 tokens (GPT-2 tokenizer with padding)
    \item \textbf{Parameters:} 124,119,552
    \item \textbf{Context length:} 512 tokens
    \item \textbf{Position encoding:} Learned position embeddings
    \item \textbf{MLP:} Asymmetric with learned PReLU activations, or GeLU as in \S\ref{sec:empirical} and Appendix \ref{app:gelu_detailed}
    \item \textbf{Training tokens:} 500M (FineWeb-Edu)
    \item \textbf{Batch size:} 8
    \item \textbf{Gradient accumulation:} 1
\end{itemize}

\subsection{Optimizer Hyperparameters}\label{app-HPs}

Table~\ref{tab:optimizer_hps} presents the hyperparameters used for each optimizer, following a period of shorter-run hyperparameter optimization.

\begin{table}[h]
\centering
\caption{Optimizer hyperparameters for main  experiments (with additional trials with different values not significantly affecting results noted parenthetically).}
\label{tab:optimizer_hps}
\begin{tabular}{llp{8cm}}
\toprule
Optimizer & Parameter & Value \\
\midrule
\multirow{4}{*}{ECD} & Learning rate ($\hat{\ell}$) & 1.0 \\
 & $\eta$ (friction coefficient) & 100 \\
 & $F_0$ & 2 \\
 & Weight decay & 0 \\
 & chaos parameter $\nu$ & 0 ~~~ (0.1)\\
\midrule
\multirow{4}{*}{SGDM} & Learning rate & 0.03 \\
 & Momentum & 0.95 \\
 & Nesterov & True \\
 & Weight decay & 0 \\
\midrule
\multirow{4}{*}{AdamW} & Learning rate & 0.0001 \\
 & $\beta_1, \beta_2$ & 0.9, 0.999 \\
 & $\epsilon$ & $10^{-8}$ \\
 & Weight decay & 0 ~~ (0.01) \\
\midrule
\multirow{5}{*}{SOAP} & Learning rate & 0.0003 \\
 & $\beta_1, \beta_2$ & 0.95, 0.95 \\
 & Precondition frequency & 10 \\
 & Max precondition dim & 10000 \\
 & Weight decay & 0.01 \\
\bottomrule
\end{tabular}
\end{table}

\section{Logic puzzles in more detail: protocol, optimizer comparison and illustrative examples}

{

\subsection{Logic Puzzle Evaluation}
\label{app:logic_puzzles}

Beyond validation loss, we evaluate whether different optimizer--symmetry configurations produce models with qualitatively different reasoning capabilities. We design a suite of logic puzzle tasks that probe pattern completion, in-context retrieval, and simple inference---capabilities that emerge in language models but are not directly measured by perplexity.

\textbf{Evaluation protocol.}
We evaluate 14 tasks across six categories: \textit{pattern\_numeric} (sequence completion, e.g., ``1, 2, 3, 4,'' $\to$ ``5''), \textit{pattern\_alpha} (alphabetic sequences including wrap-around), \textit{retrieval\_near} (recalling facts from recent context), \textit{retrieval\_far} (recalling facts from earlier context), \textit{simple\_inference} (modus ponens reasoning), and \textit{copy} (exact repetition). For each completion task, we measure whether the expected token appears in the model's top-1 or top-5 predictions. For models trained with $b_Q$ or $b_V$ biases, we apply the mean bias values ($\mathbb{E}[b_Q] = 0.5$, and $\mathbb{E}[b_V]$ matching training) at inference time.

To assess data contamination, we searched the FineWeb-Edu training corpus for our evaluation prompts. While simple counting sequences (``1, 2, 3, 4, 5'') appear in training data (42 occurrences), the majority of tasks---including Fibonacci continuation, wrap-around alphabet (X, Y, Z, A $\to$ B), all retrieval tasks, all inference tasks, and all copy tasks---are completely novel.

\subsubsection{Effect of symmetry breaking on ECD}

ECD uniformly improves with symmetry breaking on the best validation loss.  It is interesting to investigate how well this correlates with our downstream logic task.  
We analyzed four independently trained ECD models (seeds 42, 83, 123, and 789) with the $b_Q + b_V$ configuration ($\mathbb{E}[b_V] = 0.5$). 
In Table \ref{tab:ecd-fourseed-nonzerobVmean-allmetrics} we see that fully symmetry-broken (with nonzero mean biases) ECD benefits or remains neutral in 3 of 4 seeds, with one seed leading to worse logic performance, and in table \ref{tab:seed42_logic_corrected} we collect detailed results and optimizer comparisons for seed 42.

\begin{table}[H]
\centering
\caption{Summary: All Metrics Across 4 Seeds for ECD, comparing symmetric with fully symmetry broken (with nonzero mean biases $\mathbb{E}[b_Q] = \mathbb{E}[b_V] =0.5$).  Validation loss consistently improves with symmetry breaking, while logic performance improves or remains neutral in the majority of the cases.  }\label{tab:ecd-fourseed-nonzerobVmean-allmetrics}
\begin{tabular}{lcccc}
\toprule
Seed & $\Delta$ Val Loss & $\Delta$ Top-5 & $\Delta$ Top-1 & Pattern \\
\midrule
42 & $-0.56$ & $+4.1$ & $+12.5$ & Benefits across all \\
123 & $-0.57$ & $-8.3$ & $-16.6$ & Val improves, logic hurt \\
789 & $-0.41$ & $0.0$ & $0.0$ & Val improves, logic neutral \\
83 & $-0.42$ & $+4.2$ & $+16.7$ & Benefits across all \\
\bottomrule
\end{tabular}
\end{table}

It is interesting to consider how this correlates with alignment patterns.  
First, we note a commonality among all four models:  semantic coherence consistently arises.
Despite zero overlap in the specific tokens most aligned with $b_Q$ across seeds, all four models learned consistent semantic patterns. Table~\ref{tab:cross-seed-enrichment} shows the log$_2$ enrichment ratio (top-aligned vs.\ bottom-aligned tokens) for five semantic categories. 
\begin{table}[H]
\centering
\caption{Semantic category enrichment (log$_2$) across seeds. Positive values indicate enrichment in attention-enhanced tokens; negative values indicate enrichment in attention-suppressed tokens.}
\label{tab:cross-seed-enrichment}
\begin{tabular}{lcccc}
\toprule
Category & Seed 42 & Seed 83 & Seed 123 & Seed 789 \\
 & (benefit) & (benefit) & (hurt) & (neutral) \\
\midrule
Sentence starters & $+3.19$ & $+3.11$ & $+1.97$ & $+4.49$ \\
Interrogatives & $+3.19$ & $+4.90$ & $+4.09$ & $+3.79$ \\
Punctuation & $+1.42$ & $+2.10$ & $+0.12$ & $+2.26$ \\
Function words & $+1.19$ & $+2.73$ & $+0.42$ & $+1.50$ \\
\midrule
Unicode/noise & $-1.25$ & $-1.49$ & $-0.81$ & $-0.62$ \\
\bottomrule
\end{tabular}
\end{table}
\noindent All seeds enhance attention to structural markers (sentence starters, interrogatives, punctuation, function words) while suppressing unicode artifacts and encoding errors.

\begin{table}[h]
\caption{Semantic category enrichment (log$_2$) for ECD $b_Q + b_V$ PreLU models across four seeds. All seeds show consistent direction: structural markers enhanced, noise suppressed.}
\label{tab:semantic_revised}
\vskip 0.15in
\begin{center}
\begin{small}
\begin{tabular}{@{}lcccc@{}}
\toprule
\sc Category & \multicolumn{4}{c}{Seed (outcome)} \\
 & 42 & 83 & 123 & 789 \\
 & (ben.) & (ben.) & (hurt) & (neut.) \\
\midrule
\sc Sent.\ start. & $+3.19$ & $+3.11$ & $+1.97$ & $+4.49$ \\
\sc Interrog. & $+3.19$ & $+4.90$ & $+4.09$ & $+3.79$ \\
\sc Punctuation & $+1.42$ & $+2.10$ & $+0.12$ & $+2.26$ \\
\sc Func.\ words & $+1.19$ & $+2.73$ & $+0.42$ & $+1.50$ \\
\midrule
\sc Unicode/noise & $-1.25$ & $-1.49$ & $-0.81$ & $-0.62$ \\
\bottomrule
\end{tabular}
\end{small}
\end{center}
\vskip -0.1in
\end{table}

\begin{table}
\caption{Semantic category enrichment (log$_2$) for ECD symmetry breaking GELU models}
\label{tab:semantic-gelu}
\begin{tabular}{@{}lccc@{}}
\toprule
\sc Category & \multicolumn{3}{c}{bQonly Seed (outcome)} \\
 & 42 & 83 & 789 \\
 & (neut.) & (neut.) & (ben.) \\
\midrule
\sc Sent.\ start. & $+3.45$ & $+3.55$ & $+2.99$ \\
\sc Interrog. & $+5.31$ & $+1.58$ & $+3.37$ \\
\sc Punctuation & $-0.55$ & $-1.43$ & $+0.52$ \\
\sc Func.\ words & $-0.29$ & $-0.48$ & $+0.06$ \\
\midrule
\sc Unicode/noise & $+0.14$ & $-1.38$ & $-1.40$ \\
\bottomrule
\end{tabular}

\begin{tabular}{@{}lccccc@{}}
\toprule
\sc Category & \multicolumn{5}{c}{bQbV Configuration (outcome)} \\
 & s42 & V00-s83 & V00-s789 & V05-s83 & V05-s789 \\
 & (ben.) & (ben.) & (ben.) & (hurt) & (ben.) \\
\midrule
\sc Sent.\ start. & $+1.57$ & $+3.54$ & $+1.97$ & $+3.58$ & $+1.78$ \\
\sc Interrog. & $-0.29$ & $+3.54$ & $+1.88$ & $+2.32$ & $+4.39$ \\
\sc Punctuation & $+0.70$ & $-0.66$ & $+0.72$ & $-0.58$ & $+0.59$ \\
\sc Func.\ words & $-2.32$ & $-0.37$ & $-0.47$ & $-0.43$ & $+0.95$ \\
\midrule
\sc Unicode/noise & $-1.56$ & $-0.65$ & $-0.76$ & $-1.56$ & $-2.43$ \\
\bottomrule
\end{tabular}
\end{table}

However, we also find interesting distinctions among the logic-improved or neutral ECD models versus the logic-worsened seed 123 symmetry breaking model:

\begin{enumerate}[leftmargin=*]
\item \textbf{Noise suppression strength.} Seeds showing logic improvement had stronger suppression of unicode/noise tokens ($-1.25$ to $-1.49$) compared to the hurt and neutral seeds ($-0.62$ to $-0.81$). This suggests effective $b_Q$ alignment requires not only enhancing attention to structural tokens but also cleaning attention away from encoding artifacts.

\item \textbf{Punctuation enrichment.} Seed 123 (hurt on logic by symmetry breaking) had drastically weaker punctuation enrichment ($+0.12$) compared to all other seeds ($+1.42$ to $+2.26$). Punctuation tokens serve as structural boundary markers that may be particularly important for logic puzzle performance.

\item \textbf{Symmetric baseline quality.} Seed 123 had the strongest symmetric baseline performance (70.8\% Top-5) while seeds that benefited had weaker baselines (62.5--66.7\%). 
\end{enumerate}

\textbf{Layer-Specific Patterns.}
Analysis of Layer 0 revealed additional differences. Seed 123 -- the one model of the four that degraded on logic with the symmetry breaking -- exhibited extreme suppression of function words at Layer 0 (enrichment $-3.38$) compared to benefit seeds ($-0.82$ to $-1.57$). Function words carry syntactic information critical for reasoning; excessive early suppression may cause irrecoverable information loss. Furthermore, while seed 123 showed strong noise suppression at Layer 0 ($-4.20$), this pattern reversed in middle layers (enrichment $+2.32$ at Layer 6), whereas benefit seeds maintained more consistent suppression across the network.

These results suggest that the variable effect of symmetry breaking on downstream reasoning reflects not whether $b_Q$ learns semantic structure---it consistently does---but whether the specific alignment pattern complements or conflicts with representations learned through other pathways.

A similar semnatic enrichment structure appears for GELU  in Table \ref{tab:semantic-gelu}.

\subsubsection{Optimizer comparisons and task breakdown}

Table~\ref{tab:seed42_logic_corrected} presents results for all seed 42 models with Table \ref{tab:logic_task_breakdown_seed42} decomposing this into specific tasks.  Table~\ref{tab:seed789_logic} gives optimizer comparison for the seed (789) on which symmetry breaking is neutral for ECD on logic. The results reveal interactions between optimizer choice and symmetry breaking, with more nuanced results than would be predicted by best validation loss alone.  

\begin{table}[H]
\centering
\caption{Logic puzzle performance for all seed 42 models, showing a typical example where ECD improves with symmetry breaking on logic (in addition to its consistent improvement in overall validation loss). $\Delta$ shows change in Top-5 accuracy from symmetric baseline for each optimizer. Best Top-5 result(s) per optimizer in \textbf{bold}.}
\label{tab:seed42_logic_corrected}
\begin{tabular}{llccccc}
\toprule
Optimizer & Configuration & Val.\ Loss & Logic Loss & Top-5 & Top-1 & $\Delta$ Top-5 \\
\midrule
ECD  & Symmetric                       & 3.93 & 3.13 & 66.7\% & 20.8\% & -- \\
     & $b_Q$ only                      & 3.75 & 3.17 & \textbf{70.8\%} & 16.7\% & $+4.2$ \\
     & $b_Q{+}b_V$ ($\mathbb{E}[b_V]{=}0$)   & 3.36 & 3.58 & 66.7\% & 16.7\% & $+0.0$ \\
     & $b_Q{+}b_V$ ($\mathbb{E}[b_V]{=}0.5$) & 3.37 & 3.43 & \textbf{70.8\%} & \textbf{33.3\%} & $+4.2$ \\
\midrule
SGDM & Symmetric                       & 3.85 & \textbf{3.01} & \textbf{70.8\%} & \textbf{33.3\%} & -- \\
     & $b_Q$ only                      & 3.97 & 3.50 & 50.0\% & 16.7\% & $-20.8$ \\
     & $b_Q{+}b_V$ ($\mathbb{E}[b_V]{=}0$)   & 3.67 & 4.36 & 33.3\% &  0.0\% & $-37.5$ \\
     & $b_Q{+}b_V$ ($\mathbb{E}[b_V]{=}0.5$) & 3.66 & 4.10 & 58.3\% & 12.5\% & $-12.5$ \\
\midrule
AdamW & Symmetric                      & 3.38 & 3.31 & \textbf{75.0\%} & 16.7\% & -- \\
      & $b_Q$ only                     & 3.35 & 3.36 & \textbf{75.0\%} & 16.7\% & $+0.0$ \\
      & $b_Q{+}b_V$ ($\mathbb{E}[b_V]{=}0$)  & 3.53 & 3.92 & 58.3\% &  4.2\% & $-16.7$ \\
      & $b_Q{+}b_V$ ($\mathbb{E}[b_V]{=}0.5$) & 3.74 & 4.62 & 25.0\% &  4.2\% & $-50.0$ \\
\midrule
SOAP & Symmetric                       & 3.49 & 3.69 & \textbf{66.7\%} & 16.7\% & -- \\
     & $b_Q$ only                      & 3.52 & 3.59 & \textbf{66.7\%} & \textbf{25.0\%} & $+0.0$ \\
     & $b_Q{+}b_V$ ($\mathbb{E}[b_V]{=}0$)   & 3.40 & 3.43 & 58.3\% & \textbf{25.0\%} & $-8.3$ \\
     & $b_Q{+}b_V$ ($\mathbb{E}[b_V]{=}0.5$) & 3.39 & 3.49 & 62.5\% & 20.8\% & $-4.2$ \\
\bottomrule
\end{tabular}
\end{table}

\begin{table}[H]
\centering
\caption{Top-5 accuracy (\%) by task category across optimizer-symmetry configurations (seed 42). All models achieve strong performance on inference tasks; differences emerge on pattern completion and retrieval.}
\label{tab:logic_task_breakdown_seed42}
\begin{tabular}{@{}lcccccc|c@{}}
\toprule
\textbf{Configuration} & \textbf{Numeric} & \textbf{Alpha} & \textbf{Ret-Near} & \textbf{Ret-Far} & \textbf{Inference} & \textbf{Copy} & \textbf{Avg} \\
\midrule
ECD symmetric        & 50 & \textbf{100} & \textbf{100} & 50 & \textbf{100} & 0 & 66.7 \\
ECD $b_Q{+}b_V$      & 75 & \textbf{100} & \textbf{100} & 50 & \textbf{100} & 0 & 70.8 \\
\midrule
SGDM symmetric       & 75 & 50 & \textbf{100} & \textbf{100} & \textbf{100} & 0 & 70.8 \\
SGDM $b_Q{+}b_V$     & 50 & \textbf{100} & 50 & 0 & \textbf{100} & 50 & 58.3 \\
\midrule
AdamW symmetric      & 50 & 50 & 50 & \textbf{100} & \textbf{100} & \textbf{100} & \textbf{75.0} \\
SOAP $b_Q{+}b_V$     & 75 & 50 & \textbf{100} & 50 & \textbf{100} & 0 & 62.5 \\
\bottomrule
\end{tabular}
\end{table}

Several patterns emerge from this breakdown:
\begin{itemize}
    \item \textbf{Inference tasks}: All configurations achieve 100\% on simple modus ponens reasoning.
    \item \textbf{Pattern tasks}: ECD excels at alphabetic patterns (100\% for both symmetric and $b_Q{+}b_V$), while numeric patterns are more variable.
    \item \textbf{Retrieval tasks}: SGDM symmetric achieves the only 100\% on both near- and far-context retrieval; other configurations show mixed performance.
    \item \textbf{Copy tasks}: The hardest category---only AdamW symmetric achieves Top-5 success, suggesting this task probes capabilities that most configurations do not reliably develop.
\end{itemize}

\begin{table}[H]
\centering
\caption{Logic puzzle performance for seed 789 models, showing an example where symmetry breaking is neutral for ECD on this task, and where SGDM in a symmetric attention architecture does well on logic despite a worse overall validation loss. The effect of symmetry breaking varies by optimizer, consistent with seed 42 patterns.}
\label{tab:seed789_logic}
\begin{tabular}{llccccc}
\toprule
Optimizer & Configuration & Val.\ Loss & Logic Loss & Top-5 & Top-1 & $\Delta$ Top-5 \\
\midrule
ECD   & Symmetric                       & 3.81 & 3.38 & 62.5\% & 16.7\% & -- \\
      & $b_Q{+}b_V$ ($\mathbb{E}[b_V]{=}0.5$) & 3.40 & 3.30 & 62.5\% & 16.7\% & $+0.0$ \\
\midrule
SGDM  & Symmetric                       & 3.96 & 3.17 & \textbf{79.2\%} & 16.7\% & -- \\
      & $b_Q{+}b_V$ ($\mathbb{E}[b_V]{=}0.5$) & 3.65 & 4.39 & 54.2\% & 4.2\% & $-25.0$ \\
\midrule
AdamW & Symmetric                       & 3.37 & 3.18 & \textbf{75.0\%} & \textbf{29.2\%} & -- \\
      & $b_Q{+}b_V$ ($\mathbb{E}[b_V]{=}0.5$) & 3.75 & 4.44 & 50.0\% & 12.5\% & $-25.0$ \\
\midrule
SOAP  & Symmetric                       & 3.48 & 3.49 & 58.3\% & 8.3\% & -- \\
      & $b_Q{+}b_V$ ($\mathbb{E}[b_V]{=}0.5$) & 3.36 & \textbf{2.96} & \textbf{70.8\%} & \textbf{29.2\%} & $+12.5$ \\
\bottomrule
\end{tabular}
\end{table}
}


{

\subsection{Illustrative Examples of Logic Puzzle Performance}
\label{app:logic_examples}

To provide concrete insight into how symmetry breaking affects model behavior, we present detailed examples comparing model predictions across optimizers and symmetry configurations.

\subsubsection{Concrete Prediction Examples}

Tables~\ref{tab:inference_example}--\ref{tab:retrieval_example} show the actual top-5 predictions made by different models on representative tasks.

\textbf{Modus Ponens Inference.} Table~\ref{tab:inference_example} shows predictions for a simple logical inference task. All configurations successfully place the expected token in their top-5, demonstrating that basic deductive reasoning is robust across optimizer-symmetry combinations.

\begin{table}[h!]
\centering
\caption{Predictions on modus ponens inference task. Prompt: ``\textit{If it rains, the ground gets wet. It is raining. The ground}''. Expected first token: ``\texttt{ gets}''.}
\label{tab:inference_example}
\begin{tabular}{@{}l|ccccc|cc@{}}
\toprule
\textbf{Config} & \multicolumn{5}{c|}{\textbf{Top-5 Predictions}} & \textbf{Rank} & \textbf{Success} \\
\midrule
ECD $b_Q{+}b_V$ & \texttt{is} & \texttt{\textbf{gets}} & \texttt{has} & \texttt{does} & \texttt{,} & 2 & \cmark \\
ECD symmetric & \texttt{is} & \texttt{\textbf{gets}} & \texttt{will} & \texttt{can} & \texttt{has} & 2 & \cmark \\
SGDM symmetric & \texttt{is} & \texttt{\textbf{gets}} & \texttt{will} & \texttt{can} & \texttt{has} & 2 & \cmark \\
AdamW symmetric & \texttt{is} & \texttt{has} & \texttt{will} & \texttt{\textbf{gets}} & \texttt{can} & 4 & \cmark \\
\bottomrule
\end{tabular}
\end{table}

\textbf{In-Context Fact Retrieval.} Table~\ref{tab:retrieval_example} tests the model's ability to retrieve a fact (``Alice likes apples'') stated earlier in a multi-fact context.

\begin{table}[h!]
\centering
\caption{Predictions on in-context retrieval task. Prompt: ``\textit{Alice likes apples. Bob likes bananas. Carol likes cherries. Dave likes dates. Alice likes}''. Expected: ``\textit{ apples}''.}
\label{tab:retrieval_example}
\begin{tabular}{@{}l|ccccc|cc@{}}
\toprule
\textbf{Config} & \multicolumn{5}{c|}{\textbf{Top-5 Predictions}} & \textbf{Rank} & \textbf{Success} \\
\midrule
ECD $b_Q{+}b_V$ & \texttt{\textbf{apples}} & \texttt{to} & \texttt{the} & \texttt{a} & \texttt{fruit} & 1 & \cmark \\
ECD symmetric & \texttt{\textbf{apples}} & \texttt{the} & \texttt{.} & \texttt{to} & \texttt{orange} & 1 & \cmark \\
SGDM symmetric & \texttt{\textbf{apples}} & \texttt{to} & \texttt{her} & \texttt{the} & \texttt{fruit} & 1 & \cmark \\
AdamW symmetric & \texttt{\textbf{apples}} & \texttt{the} & \texttt{a} & \texttt{to} & \texttt{bananas} & 1 & \cmark \\
\bottomrule
\end{tabular}
\end{table}

\textbf{Near-Context Retrieval.} Table~\ref{tab:near_retrieval_example} shows a retrieval task with closer context. ECD $b_Q{+}b_V$ and SGDM symmetric achieve rank 1 while other configurations succeed at lower ranks, illustrating how symmetry breaking can sharpen attention to relevant context.

\begin{table}[h!]
\centering
\caption{Predictions on near-context retrieval. Prompt: ``\textit{The color is blue. The shape is round. The color is}''. Expected: ``\textit{ blue}''.}
\label{tab:near_retrieval_example}
\begin{tabular}{@{}l|ccccc|cc@{}}
\toprule
\textbf{Config} & \multicolumn{5}{c|}{\textbf{Top-5 Predictions}} & \textbf{Rank} & \textbf{Success} \\
\midrule
ECD $b_Q{+}b_V$ & \texttt{\textbf{blue}} & \texttt{red} & \texttt{white} & \texttt{black} & \texttt{yellow} & 1 & \cmark \\
ECD symmetric & \texttt{green} & \texttt{black} & \texttt{white} & \texttt{\textbf{blue}} & \texttt{yellow} & 4 & \cmark \\
SGDM symmetric & \texttt{\textbf{blue}} & \texttt{yellow} & \texttt{white} & \texttt{black} & \texttt{green} & 1 & \cmark \\
AdamW symmetric & \texttt{white} & \texttt{black} & \texttt{\textbf{blue}} & \texttt{yellow} & \texttt{red} & 3 & \cmark \\
\bottomrule
\end{tabular}
\end{table}

\textbf{Wrap-Around Alphabetic Sequence.} Table~\ref{tab:alpha_example} tests whether the model can continue an alphabetic sequence that wraps from Z to A to B. 

\begin{table}[h!]
\centering
\caption{Predictions on wrap-around alphabetic sequence. Prompt: ``\textit{X, Y, Z, A,}''. Expected: ``\textit{ B}''.}
\label{tab:alpha_example}
\begin{tabular}{@{}l|ccccc|cc@{}}
\toprule
\textbf{Config} & \multicolumn{5}{c|}{\textbf{Top-5 Predictions}} & \textbf{Rank} & \textbf{Success} \\
\midrule
ECD $b_Q{+}b_V$ & \texttt{and} & \texttt{\textbf{B}} & \texttt{et} & \texttt{K} & \texttt{D} & 2 & \cmark \\
ECD symmetric & \texttt{Z} & \texttt{\textbf{B}} & \texttt{C} & \texttt{A} & \texttt{E} & 2 & \cmark \\
SGDM $b_Q{+}b_V$ & \texttt{\textbf{B}} & \texttt{K} & \texttt{and} & \texttt{Y} & \texttt{Z} & 1 & \cmark \\
AdamW symmetric & \texttt{Z} & \texttt{K} & \texttt{and} & \texttt{G} & \texttt{Y} & 8 & \xmark \\
\bottomrule
\end{tabular}
\end{table}

{

\section{Comparison with Learned Attention Biases}
\label{app:learned_biases}

Our main protocol uses \emph{unlearned} biases $b_Q$ and $b_V$ with compomponents that are resampled from $\mathcal{N}(\mu, \sigma^2 )$ each batch as described in \S\ref{sec-rotation-breaking-bQ-bV}. A natural question is whether making these biases learnable (as \texttt{nn.Parameter}) would improve performance.

We compare:
\begin{enumerate}
    \item \textbf{Unlearned (main protocol)}: $b_Q, b_V$ resampled each batch, not trained
    \item \textbf{Learned, nonzero init}: $b_Q, b_V$ as trainable parameters, initialized from $\mathcal{N}(0.5, \sigma^2)$
\end{enumerate}

\subsubsection{Results}

Table~\ref{tab:learned_vs_unlearned} compares validation loss and logic puzzle performance between learned and unlearned biases. For unlearned biases, we report the $b_Q{+}b_V$ configuration with $\mathbb{E}[b_V] = 0.5$, which achieves the best results for ECD (see Table~\ref{tab:seed42_logic_corrected}).

\begin{table}[h!]
\centering
\caption{Comparison of learned vs.\ unlearned attention biases. Learned biases achieve comparable or better validation loss and logic puzzle performance for most optimizers. Adam with learned biases achieves the best overall logic loss (3.15), while ECD with unlearned biases achieves the highest Top-1 accuracy (33.3\%).}
\label{tab:learned_vs_unlearned}
\begin{tabular}{@{}l|cccc|cccc@{}}
\toprule
& \multicolumn{4}{c|}{\textbf{Unlearned} ($b_Q{+}b_V$, $\mathbb{E}[b_V]{=}0.5$)} & \multicolumn{4}{c}{\textbf{Learned} (nonzero init)} \\
\textbf{Optimizer} & Val.\ Loss & Logic Loss & Top-5 & Top-1 & Val.\ Loss & Logic Loss & Top-5 & Top-1 \\
\midrule
ECD   & 3.37 & 3.43 & \textbf{70.8\%} & \textbf{33.3\%} & 3.36 & 3.51 & 64.3\% & 7.1\% \\
SOAP  & 3.39 & 3.49 & 62.5\% & 20.8\% & 3.38 & 3.67 & 64.3\% & 21.4\% \\
SGDM  & 3.66 & 4.10 & 58.3\% & 12.5\% & 3.30 & 3.41 & 57.1\% & 21.4\% \\
AdamW & 3.74 & 4.62 & 25.0\% & 4.2\% & \textbf{3.00} & \textbf{3.15} & 64.3\% & 21.4\% \\
\bottomrule
\end{tabular}
\end{table}

Several patterns emerge:

\textbf{Learned biases improve Adam substantially.} Adam with learned biases achieves both the best validation loss (3.00) and best logic loss (3.15) across all configurations tested. This represents a large improvement over Adam with unlearned biases (val.\ loss 3.74, logic loss 4.62).  The learned bias case has rotational symmetry in the attention heads, but Adam breaks it with its preferred basis.  

\textbf{Top-5 accuracy is comparable.} Learned and unlearned biases achieve similar Top-5 accuracy (57--64\% vs.\ 58--71\%), indicating that the models develop comparable reasoning capabilities at the coarse-grained level. The main difference appears in Top-1 accuracy, where ECD with unlearned biases uniquely achieves 33.3\% (vs.\ 7.1\% with learned biases).

\textbf{ECD benefits most from unlearned biases for Top-1 accuracy.}  ECD's Top-5 accuracy is somewhat worse with learned (64.3\%) as opposed to unlearned (70.8\%) biases, its Top-1 accuracy drops substantially (33.3\% $\to$ 7.1\%) for learned biases. This fits with the theoretical predictions that symmetry breaking and chaotic dynamics improve ECD performance.  

\textbf{SGDM and SOAP show mixed results.} Both SGDM and SOAP achieve slightly better logic loss with unlearned biases for SOAP (3.49 vs.\ 3.67), while SGDM improves with learned biases (3.41 vs.\ 4.10). Top-5 accuracy remains similar in both cases.

\textbf{Zero-initialization also performs well.} We also tested learned biases initialized to zero rather than $\mathcal{N}(0.5, \sigma^2)$. Despite developing much smaller bias magnitudes (total $\|b_Q\| \approx 6$--$42$ vs.\ $\sim$165 for nonzero-init), zero-init models can achieve competitive results; in particular Adam zero-init achieves the lowest logic loss overall (2.64), suggesting that the model can effectively learn to utilize even small learned biases.

\subsubsection{Discussion}

For the optimizers with comparable performance between learned and unlearned biases, the primary benefit of attention biases may come from their \emph{presence} rather than their \emph{stochasticity}.
The models can in principle learn the entire mechanism of key alignment with query biases, learning to grow or maintain the useful direction $b_Q$ (depending on initialization).

However, the advantage of unlearned biases for ECD's Top-1 accuracy points to a more nuanced picture.  
One interpretation, following our motivation in \S\ref{sec-ECD-SGDM-conserved-angular-momenta}, is that unlearned biases preserve a form of continuous symmetry breaking that benefits ECD's Hamiltonian dynamics. With learned biases, the model can ``rotate with'' the bias direction during training, potentially reducing the effective symmetry breaking. With unlearned biases, there is a mean that the model can align helpful tokens with, while the precise direction varies each batch, forcing the model to learn representations that are robust.  

\subsubsection{Practical Recommendation}

For practitioners optimizing primarily for validation loss, learned biases with Adam provide the best results (val.\ loss 3.00). For applications requiring high-confidence reasoning (Top-1 accuracy), unlearned biases with ECD remain the best choice. When Top-5 accuracy suffices, either protocol achieves comparable results, and ECD provides a simpler, $2N$-variable theoretically predictive option as discussed in the main text.  

}
}



\section{GELU examples: Detailed Results}
\label{app:gelu_detailed}

This appendix provides detailed results for the GELU versions summarized in Section~\ref{sec:empirical}.

\subsection{Alignment Strength}
\label{app:gelu_alignment}

Table~\ref{tab:gelu_alignment} shows $W_K$ alignment with $\mathbb{E}[b_Q]$ for all symmetry-broken GELU configurations. All achieve alignment far above the random threshold of 0.58, though slightly weaker than PReLU models.

\begin{table}[h]
\caption{Alignment strength for GELU ECD models (symmetry-broken configurations).}
\label{tab:gelu_alignment}
\centering
\small
\begin{tabular}{lccc}
\toprule
\textbf{Configuration} & \textbf{Mean Top} & \textbf{\% $>$ 0.58} & \textbf{Max} \\
\midrule
bQonly-seed42 & 0.701 & 91.2\% & 0.921 \\
bQonly-seed83 & 0.706 & 85.5\% & 0.892 \\
bQonly-seed789 & 0.721 & 90.2\% & 0.890 \\
bQbV-seed42 & 0.695 & 85.6\% & 0.900 \\
bQbV-V00-seed83 & 0.708 & 89.9\% & 0.907 \\
bQbV-V00-seed789 & 0.712 & 88.1\% & 0.889 \\
bQbV-V05-seed83 & 0.709 & 88.7\% & 0.915 \\
bQbV-V05-seed789 & 0.696 & 85.5\% & 0.881 \\
\midrule
\textbf{GELU Mean} & \textbf{0.706} & \textbf{88.1\%} & -- \\
\textbf{PReLU Mean} & \textbf{0.731} & \textbf{97.2\%} & -- \\
\bottomrule
\end{tabular}
\end{table}

\subsection{Validation Loss}
\label{app:gelu_val_loss}

Table~\ref{tab:gelu_val_loss} compares validation loss between symmetric and symmetry-broken configurations. GELU achieves much lower absolute validation loss than PReLU, but the improvement from symmetry breaking is correspondingly smaller.

\begin{table}[h]
\caption{Validation loss comparison: GELU vs PReLU (ECD optimizer).}
\label{tab:gelu_val_loss}
\centering
\small
\begin{tabular}{lcccccc}
\toprule
& \multicolumn{3}{c}{\textbf{PReLU}} & \multicolumn{3}{c}{\textbf{GELU}} \\
\cmidrule(lr){2-4} \cmidrule(lr){5-7}
\textbf{Seed} & Sym & SB & $\Delta$ & Sym & Best SB & $\Delta$ \\
\midrule
42  & 3.93 & 3.37 & $-0.56$ & 3.185 & 3.137 & $-0.049$ \\
83  & 3.82 & 3.40 & $-0.42$ & 3.141 & 3.128 & $-0.013$ \\
789 & 3.81 & 3.40 & $-0.41$ & 3.176 & 3.152 & $-0.025$ \\
\midrule
\textbf{Mean} & 3.88 & 3.39 & $\mathbf{-0.49}$ & 3.17 & 3.14 & $\mathbf{-0.03}$ \\
\bottomrule
\end{tabular}
\vspace{1mm}
\begin{flushleft}
\small\textit{Note:} PReLU includes seed 123 in paper averages (3.94/3.37). GELU ``Best SB'' is the best-performing symmetry-broken configuration for each seed.
\end{flushleft}
\end{table}

\subsection{Logic Puzzle Performance}
\label{app:gelu_logic}

Table~\ref{tab:gelu_logic} shows logic puzzle performance for all ECD GELU configurations. Despite smaller validation loss improvements, symmetry breaking benefits logic puzzles more consistently for GELU than PReLU.

\begin{table}[h]
\caption{Logic puzzle performance for GELU ECD models.}
\label{tab:gelu_logic}
\centering
\small
\begin{tabular}{llcccc}
\toprule
\textbf{Seed} & \textbf{Config} & \textbf{Logic Loss} & \textbf{Top-5} & \textbf{$\Delta$ Top-5} & \textbf{Effect} \\
\midrule
42 & symmetric & 2.837 & 71.4\% & -- & -- \\
42 & bQonly & 3.003 & 71.4\% & $+0.0$ & Neutral \\
42 & bQbV-V00 & 3.095 & 78.6\% & $+7.1$ & Benefit \\
\midrule
83 & symmetric & 2.710 & 78.6\% & -- & -- \\
83 & bQonly & 2.913 & 78.6\% & $+0.0$ & Neutral \\
83 & bQbV-V00 & 2.817 & 85.7\% & $+7.1$ & Benefit \\
83 & bQbV-V05 & 3.362 & 71.4\% & $-7.1$ & Hurt \\
\midrule
789 & symmetric & 3.111 & 64.3\% & -- & -- \\
789 & bQonly & 2.822 & 78.6\% & $+14.3$ & Benefit \\
789 & bQbV-V00 & 2.820 & 85.7\% & $+21.4$ & Benefit \\
789 & bQbV-V05 & 2.956 & 78.6\% & $+14.3$ & Benefit \\
\bottomrule
\end{tabular}
\vspace{1mm}
\begin{flushleft}
\small\textit{Effect classification:} Benefit ($\Delta > +2$), Neutral ($|\Delta| \leq 2$), Hurt ($\Delta < -2$). \\
\small\textit{Summary:} 5/8 Benefit (62.5\%), 2/8 Neutral (25\%), 1/8 Hurt (12.5\%). \textbf{Not hurt: 7/8 (87.5\%).}
\end{flushleft}
\end{table}

\subsection{Semantic alignment example}

See table \ref{tab:punct_example} for a case illustrating the alignment effect enhancing punctuation and suppressing unicode, in a symmetry-breaking-improved GELU model.  

BPE tokenizers trained on web-scraped data inherit vocabulary entries for encoding errors (the replacement character U+FFFD), partial multilingual sequences, and HTML/JavaScript fragments like \texttt{usercontent}. That our models consistently learn to suppress attention to such tokens (11/12 models, $p = 0.003$) suggests the $b_Q$ alignment mechanism provides a channel for filtering noise from the attention computation.

\begin{table}[t]
\caption{Punctuation enhancement in Layer~0, Head~10 of ECD GELU ($b_Q{+}b_V$, $\mu_V{=}0$, seed 789). This configuration achieved the largest logic improvement: $\Delta$\,Top-5 $= +21.4$ percentage points (from 64.3\% to 85.7\%). Seven of ten top-aligned tokens are punctuation; bottom-aligned tokens include encoding errors and web artifacts.}
\label{tab:punct_example}
\centering
\small
\begin{tabular}{lclc}
\toprule
\multicolumn{2}{c}{\textbf{Top Aligned}} & \multicolumn{2}{c}{\textbf{Bottom Aligned}} \\
\textbf{Token} & \textbf{Sim.} & \textbf{Token} & \textbf{Sim.} \\
\midrule
\texttt{.}       & $+0.80$ & JSON fragment        & $-0.66$ \\
\texttt{(}       & $+0.80$ & escaped backslashes  & $-0.66$ \\
\texttt{?}       & $+0.79$ & CJK + U+FFFD         & $-0.62$ \\
\texttt{/}       & $+0.78$ & \texttt{Prosecutors} & $-0.58$ \\
\texttt{!}       & $+0.77$ & \texttt{tube}        & $-0.58$ \\
\texttt{"}       & $+0.76$ & \texttt{rookies}     & $-0.57$ \\
\texttt{).}      & $+0.76$ & \texttt{cloneembed...} & $-0.57$ \\
en-dash          & $+0.75$ & \texttt{usercontent} & $-0.57$ \\
em-dash          & $+0.75$ & \texttt{acly}        & $-0.57$ \\
\bottomrule
\end{tabular}
\end{table}

\subsection{Other Optimizers}
\label{app:gelu_other_opt}

Table~\ref{tab:gelu_other_opt} shows results for Adam, SGDM, and SOAP at seed 42. Only Adam improves validation loss with symmetry breaking, yet Adam is hurt on logic puzzles. SOAP shows the opposite pattern: neutral validation loss but improved logic puzzles.  These results fit with the semantic alignment patterns as discussed in the main text.  

\begin{table}[h]
\caption{Other optimizers with GELU (seed 42).}
\label{tab:gelu_other_opt}
\centering
\small
\begin{tabular}{lccccc}
\toprule
\textbf{Optimizer} & \textbf{Val Loss $\Delta$} & \textbf{Alignment} & \textbf{Top-5 Sym} & \textbf{Top-5 SB} & \textbf{$\Delta$ Top-5} \\
\midrule
Adam & $-0.026$ & 0.66 & 85.7\% & 78.6\% & $-7.1$ (Hurt) \\
SGDM & $+0.011$ & 0.48 & 64.3\% & 50.0\% & $-14.3$ (Hurt) \\
SOAP & $+0.001$ & 0.28 & 57.1\% & 64.3\% & $+7.1$ (Benefit) \\
\midrule
ECD & $-0.025$ & 0.70 & 71.4\% & 78.6\% & $+7.1$ (Benefit) \\
\bottomrule
\end{tabular}
\vspace{1mm}
\begin{flushleft}
\small\textit{Note:} ECD row shows seed 42 only for comparison. Alignment is mean top alignment for the symmetry-broken model.
\end{flushleft}
\end{table}

\subsection{Summary}
\label{app:gelu_summary}

The GELU ablation reveals several key findings:

\begin{enumerate}[leftmargin=*,itemsep=2pt]
    \item \textbf{Alignment is architecture-general:} GELU models learn significant $W_K$ alignment with $\mathbb{E}[b_Q]$ (mean 0.71, 88\% of heads above threshold), though weaker than PReLU (0.73, 97\%).
    \item \textbf{Validation loss improvement is smaller:} Symmetry breaking improves GELU validation loss by only $-0.03$ on average, compared to $-0.49$ for PReLU. This suggests GELU's symmetric activation already captures much of the relevant structure.
    \item \textbf{Symmetry breaking is low-risk:} 87.5\% of GELU configurations (7/8) trained by ECD either benefit or are neutral on logic puzzles, with only 12.5\% hurt. This compares to 75\% not hurt for PReLU (3/4). For practitioners, symmetry breaking can be adopted with low risk of harming reasoning capabilities.
    \item \textbf{Optimal componentwise $b_V$ mean $\mu_V$ differs:} For GELU, $\mu_V= 0$ consistently outperforms $\mu_V = 0.5$ on logic puzzles, the opposite of PReLU where $\mu_V = 0.5$ was standard.
    \item \textbf{Optimizer effects vary but correlates with semantic alignment patterns:} With GELU, ECD and SOAP benefit from symmetry breaking on logic puzzles, while Adam and SGDM are hurt. Alignment strength does not reliably predict downstream benefit, but alignment semantics does predict the benefit as discussed in the main text.
\end{enumerate}


\medskip

\end{document}